\documentclass[twoside,dvipdfmx]{article}
\usepackage{ecj,palatino,epsfig,latexsym,natbib}

\usepackage{amsmath,amssymb,mathtools,stmaryrd,bm}
\usepackage{subcaption} 
\usepackage{xcolor}
\usepackage{ifthen}
\usepackage[english]{babel}

\graphicspath{{./fig/}}

\newcommand{\markupdraft}[2]{
    \ifthenelse{\equal{#1}{display}}{#2}{}
    \ifthenelse{\equal{#1}{color}}{\color{#2}}{}
}
\newcommand{\notecolored}[3][]{\markupdraft{display}{{\color{#2}\noindent[Note (#1): #3]}}}
\newcommand{\newcolored}[3][]{{\markupdraft{color}{#2}#3}
    \ifthenelse{\equal{#1}{}}{}{\markupdraft{display}{{\color{yellow!70!black}[#1]}}}} 

\newcommand{\del}[2][]{{\markupdraft{display}{{\color{orange}[removed: ``#2''[#1]]}}}} 
\newcommand{\new}[2][]{\newcolored[#1]{blue}{#2}}
\newcommand{\nnew}[2][]{\newcolored[#1]{red}{#2}}
\newcommand{\note}[2][]{\notecolored[#1]{green}{#2}}    

\renewcommand{\del}[2]{} 
\renewcommand{\markupdraft}[2]{}
\newcommand{\yohe}[1]{\note[Youhei]{\color{cyan} #1}}
\newcommand{\shin}[1]{\note[Shinichi]{\color{magenta} #1}}

\newcommand{\R}{\mathbb{R}} 
\newcommand{\T}{\mathrm{T}} 
\renewcommand{\geq}{\geqslant} 
\renewcommand{\leq}{\leqslant} 
\renewcommand{\epsilon}{\varepsilon} 
\renewcommand{\phi}{\varphi} 
\DeclareMathOperator*{\Var}{Var} 
\DeclareMathOperator*{\Cov}{Cov} 
\newcommand{\rmd}{\mathrm{d}}
\newcommand{\ind}[1]{\mathbb{I}\{#1\}}
\DeclarePairedDelimiterX{\inner}[2]{\langle}{\rangle}{#1, #2}

\newcommand{\pthetat}{p_{{\theta^{(t)}}}}
\newcommand{\pthetaj}{p_{{\theta^{(j)}}}}
\newcommand{\pthetak}{p_{{\theta^{(k)}}}}

\newcommand{\mueff}{\mu_{\mathrm{eff}}} 
\DeclareMathOperator{\rk}{rk} 
\newcommand{\mm}{m}
\newcommand{\CC}{C}
\providecommand{\pc}{p_c}
\providecommand{\cc}{c_c}
\providecommand{\cm}{c_m}
\providecommand{\cone}{c_1}
\providecommand{\cmu}{c_{\mu}}

\newcommand{\eye}{I}

\begin{document}



\title{\bf Sample Reuse via Importance Sampling in Information Geometric Optimization}  

\author{
        \name{\bf Shinichi Shirakawa} \hfill \addr{shirakawa-shinichi-bg@ynu.ac.jp}\\ 
        \addr{Faculty of Environment and Information Sciences, Yokohama National University, Kanagawa, Japan}
\AND
       \name{\bf Youhei Akimoto} \hfill \addr{akimoto@cs.tsukuba.ac.jp}\\
        \addr{Faculty of Engineering, Information and Systems, University of Tsukuba, Ibaraki, Japan}
\AND
       \name{\bf Kazuki Ouchi} \hfill \addr{kzk0520.cb@gmail.com}\\
        \addr{Graduate School of Science and Engineering, Aoyama Gakuin University, Kanagawa, Japan}
\AND
       \name{\bf Kouzou Ohara} \hfill \addr{ohara@it.aoyama.ac.jp}\\
        \addr{College of Science and Engineering, Aoyama Gakuin University, Kanagawa, Japan}
}

\maketitle

\begin{abstract}
\note{The abstract goes here.  It should be about 200 words and give the
reader a summary of the main contributions of the paper.
Remember that readers may decide to read or not to read your
paper based on what is in the abstract.  The abstract never
contains references.}
\new{In this paper we propose a technique to reduce the number of function evaluations, which is often the bottleneck of the black-box optimization, in the}\del{The}{} information geometric optimization (IGO) \new{that} is a generic framework of the probability model\nnew{-}based black-box optimization algorithms and generalizes several well-known evolutionary algorithms, such as the population\nnew{-}based incremental learning (PBIL) and the pure rank-$\mu$ update covariance matrix adaptation evolution strategy (CMA-ES). In each iteration, the IGO algorithms update the parameters of the probability distribution to the natural gradient direction estimated by Monte-Carlo with the samples drawn from the current distribution. \del{In this paper, we propose a sample reuse technique in the IGO}{}\new{Our strategy is to reuse previously generated and evaluated samples} based on the importance sampling\new{. It is a technique to reduce the estimation variance without introducing a bias in Monte-Carlo estimation.}\del{ to reduce the estimation variance. The proposed technique reuses the past samples drawn from the different distributions, employing the importance sampling twice: one in the estimate of utility value for each candidate, the other in the natural gradient estimate.}{} We apply the sample reuse technique to the PBIL and the pure rank-$\mu$ update CMA-ES and empirically investigate its effect. The experimental results show that the sample reuse helps to reduce the number of function evaluations on many benchmark functions for both the PBIL and the pure rank-$\mu$ update CMA-ES. \new{Moreover, we demonstrate how to combine the importance sampling technique with a variant of the CMA-ES \nnew{involving} an algorithmic component that is not derived in the IGO framework. }
\end{abstract}


\providecommand{\bbX}{\mathbb{X}}

\section{Introduction}

\yohe{Situations where IS may be useful in CMA
  \begin{itemize}
  \item noisy function (sigma not changing, but a very good function value obtained by chance may affect the performance drastically)
  \item asynchronous implementation (correcting samples)
  \item scenario where we would like to inject solutions (cf. Niko's injection)
  \end{itemize}
}

\paragraph{Motivation}
In the black-box optimization settings, it is often the case that the evaluation of each candidate solution is the bottleneck of the computational time for the optimization process. The objective of \nnew{a} search algorithm is to find a high-quality solution with the least number of evaluations of candidate solutions. The quality of solutions and the speed of the search are generally contradicting: a fast converging algorithm tends to be trapped by a local \nnew{minima} close to the initial search point or tends to exhibit premature convergence. \nnew{An} algorithm designer tries to realize a reasonable \nnew{trade}-off between the speed of the search and the quality of the solution. Our motivation is to construct a mechanism to reduce the number of required function evaluations of the existing and promising algorithms without deteriorating the quality of the solution. 

\paragraph{Information Geometric Optimization Framework}

The baseline algorithms of this study are the information geometric optimization (IGO) algorithms \citep{Ollivier2017}. The IGO framework is a generic framework of the probability model-based search algorithms for an arbitrary domain. It takes a parametric family of probability distributions on the search space as an input, from which the candidate solutions are generated. Then, the IGO framework provides a\nnew{n} iterative way to update the parameters of the family of probability distributions. Differently from the standard estimation of distribution algorithms (EDAs) that also maintain a family of probability distributions, the IGO algorithm does not maintain the population, i.e., the set of candidate solutions, but it maintains the parameters of a probability distribution and all information is stored in the parameters.

The IGO algorithms repeat\del{s}{} the following steps until a termination condition is satisfied. At the beginning of the search, the parameters of the probability distribution \del{is}{}\nnew{are} initialized with user-provided values. At each iteration, multiple candidate solutions, i.e., population, are generated from the probability distribution. \nnew{These} are then evaluated on the objective function. The objective function values are transformed to the utility values by a ranking-based transformation. The parameters of the probability distribution \del{is}{}\nnew{are} then updated so as to increase the expected utility values under the probability distribution. To do so, we take the so-called natural gradient \citep{Amari1998} of a function on the space of the distribution parameters. That is, we treat the expected utility as a function of the distribution parameters and take the steepest ascent step with respect to the Fisher metric. The natural gradient is estimated by Monte-Carlo using the current candidate solutions.

It is known that the IGO framework recovers some existing algorithms designed independently of the IGO framework. \citet{Akimoto2010} and \citet{Glasmachers2010} reveal that the state-of-the-art black-box randomized algorithm on \nnew{a} continuous domain, namely, the covariance matrix adaptation evolution strategy (CMA-ES), contains the components derived from the IGO framework. More precisely, a simplified variant of the CMA-ES that performs the weighted recombination and the rank-$\mu$ covariance matrix update is derived from the IGO framework. Moreover, the population\nnew{-}based incremental learning (PBIL) \citep{Baluja1994} and the compact genetic algorithm (cGA) \citep{Harik1999} for binary optimization can be derived from the IGO framework.

\paragraph{Sample Reuse}
The objective of this work is to accelerate the IGO framework without changing the working principle. 
Our strategy is to \emph{reuse} candidate solutions generated in the past iterations.

In evolutionary computation community, it is quite natural to reuse candidate solutions. In elitist strategies such as ($\mu+\lambda$)-EA, the best $\mu$ candidate solutions among $\mu$ parental (past) candidate solutions and $\lambda$ (new) candidate solutions are selected. \nnew{This} implies that some of the past candidate solutions are reused several times. The elitist strategies are still common in evolutionary computation communities especially in discrete domain, however, they are not used in the PBIL and the standard CMA-ES. A main reason of not using the past candidate solutions in the PBIL and the CMA-ES is that the update of the distribution parameters, such as the probability vector in the PBIL and the mean vector and the covariance matrix in the CMA-ES, assumes that the candidate solutions are generated from the current distribution. Using the candidate solutions from past iterations results in introducing an undesired bias in the parameter update.

\citet{Yi2009a,Yi2009} \del{has}{}\nnew{have} introduced the idea of \emph{importance mixing} into a variant of evolution strategies, namely the natural evolution strategy \citep{Wierstra2008}. It consists of two steps. In the first step, some candidate solutions from the previous iteration \del{is}{}\nnew{are} accepted according to the probability defined by the likelihood ratio at the candidate solution given the previous and the current distributions. In the second step, new candidate solutions are generated from the current distribution, and they are rejected with the probability defined by the likelihood ratio. Then, the solutions accepted in the first step and the second step \nnew{have}\del{has}{} the same likelihood values as they are evaluated on the current distribution. It is by default designed to reduce the number of evaluations per iteration. Importance mixing is effective when the objective function is unimodal as we also confirm in our experiments; however, it tends to lead to premature convergence at a local minimum when solving a multimodal function. \yohe{I do not fully understand why it happens. Statistically, the reason may be because the samples are correlated over iterations.}

\paragraph{Contribution}

We propose a sample reuse \del{mechanism}{}\nnew{technique} for the information geometric optimization framework. Our strategy is based on the \emph{importance sampling} (e.g., Chapter~4 of \citep{FishmanBook1996}), a technique to estimate the expectation of a quantity under a probability distribution via Monte-Carlo sampling with a different probability distribution. Since the IGO algorithm estimates the natural gradient of the expected utility by Monte-Carlo, the importance sampling can be applied. We introduce the importance sampling to reuse the past samples without introducing the bias and improve the accuracy of the natural gradient estimate. In the IGO framework, we employ the importance sampling twice: one in the estimate of a utility value for each candidate, the other in the natural gradient estimate. As example applications, we apply the proposed framework to PBIL and CMA-ES, where the former fully fits in the IGO framework \nnew{while} the latter contains some components that are not described in the IGO framework. This paper extends the previous work \citep{Shirakawa2015} of the sample reuse strategy for the CMA-ES. In this paper the treatment of the utility function is improved so that tie candidates can be treated formally. This allows us to extend the framework to discrete domain, where ties often occur. We investigate the effectiveness of the sample reuse mechanism both for the CMA-ES and the PBIL. 

\paragraph{Paper Organization} In Section~\ref{sec:algo} we briefly review the IGO, PBIL, and CMA-ES, \nnew{along with} the importance mixing technique. We introduce the importance \del{mixing}{}\nnew{sampling} into the IGO framework in Section~\ref{sec:reuse-igo}. Applications of the importance sampling technique to PBIL and CMA-ES are discussed in Section~\ref{sec:app}. Experimental results are discussed in Section~\ref{sec:exppbil} for PBIL and in Section~\ref{sec:expcma} for CMA-ES variants. We conclude the paper in Section~\ref{sec:conclusion}.

\section{Related Algorithms}
\label{sec:algo}

In the following, let $\bbX$ be an arbitrary search space and $f:\bbX \to \R$ be the objective function on $\bbX$ taking a value in $\R$. Without loss of generality\del{(w.l.g.)}{}, we assume that $f$ is to be minimized. 

We first introduce our baseline framework, information geometric optimization (IGO). Then, we derive the population\nnew{-}based incremental learning (PBIL) and the covariance matrix adaptation evolution strategy (CMA-ES) as instantiations of the IGO. 

\subsection{Information Geometric Optimization}
\label{sec:igo}

The information geometric optimization is a framework of the probability model-based algorithms on an arbitrary search space $\bbX$. It takes the model of probability distributions, $P_\theta$, on the search space $\bbX$ parameterized by $\theta \in \Theta \subseteq \R^D$, where $D$ is the number of parameters and maintains the parameter $\theta$ so that the probability distribution $P_\theta$ tends to concentrate at the optimal solution of $f$. It repeats the following steps:
\begin{enumerate}
\item sample $\lambda$ ($> 1$) independent candidate solutions, $x_1, \dots, x_\lambda$, from $P_\theta$;
\item evaluate the objective function value $f(x_i)$ for each $x_i$ ($i = 1, \dots, \lambda$);
\item update the parameter $\theta$ using the candidate solutions. 
\end{enumerate} 
The update step, the third step above, employs the so-called \emph{natural gradient} of a function defined over the parameter space $\Theta$ of the probability distributions equipped with the Fisher metric. We explain in the following how the parameter of the probability distribution is updated.
To set the idea, we suppose that we have full access on $f$. This assumption is later removed when algorithm instances are derived. 

The IGO transforms the original minimization problem of $f: \bbX \to \R$ to the maximization of a function $J_{\theta^{(t)}}: \Theta \to \R$ on the domain of the parameter of the probability distributions at each iteration. The function $J_{\theta^{(t)}}$ depends on the parameter $\theta^{(t)}$ at each iteration. The function value $J_{\theta^{(t)}}(\theta)$ at $\theta \in \Theta$ is defined as the expectation of the utility $W^f_{\theta^{(t)}}(x)$ over $x \sim P_\theta$, namely
\begin{equation*}
J_{\theta^{(t)}}(\theta) = \int W^f_{\theta^{(t)}}(x) p_{\theta}(x) \rmd x,
\end{equation*}
where $\rmd x$ is a reference measure on $\bbX$ and $p_{\theta}$ \nnew{is} the Radon-Nikodym derivative of $P_\theta$ with respect to $\rmd x$. The function $p_\theta$ is a probability density function on a continuous domain if $\rmd x$ is the Lebesgue measure, and it is a probability mass function on a discrete domain if $\rmd x$ is the counting measure. 

Simple examples of the definition of utility functions are $-f(x)$, $\exp(-f(x))$, etc. In the IGO framework, the utility function is a nonlinear and non-increasing transformation of the objective function to achieve the invariance to strictly increasing transformation of the objective function. The utility function, $W^f_{\theta^{(t)}}(x)$, is defined by using the quantile of $f(x)$ under $x \sim p_{\theta^{(t)}}$. Let 
\begin{equation}
q^{\leq}_{\theta^{(t)}}(x) = P_{\theta^{(t)}}[y: f(y) \leq f(x)]  \qquad \text{and} \qquad
q^{<}_{\theta^{(t)}}(x) = P_{\theta^{(t)}}[y: f(y) < f(x)] \enspace,
\label{eq:quantile}
\end{equation}
be, respectively, the probability of sampling an equal or better, and strictly better candidate solution from $P_{\theta^{(t)}}$ than a given $x \in \bbX$. Let $w: [0,1] \to \R$ be a non-increasing function. The utility value at $x$ given $\theta^{(t)}$ is then defined as follows
\begin{equation}
W^f_{\theta^{(t)}}(x) =
	\begin{cases}
	 w(q_{\theta^{(t)}}^{\leq}(x)) & (\mathrm{if} \;\; q_{\theta^{(t)}}^{\leq}(x) = q_{\theta^{(t)}}^{<}(x)) \\
	 \frac{1}{q_{\theta^{(t)}}^{\leq}(x) - q_{\theta^{(t)}}^{<}(x)} \int_{q_{\theta^{(t)}}^{<}(x)}^{q_{\theta^{(t)}}^{\leq}(x)} w(q) \rmd q  & (\mathrm{otherwise}) \enspace.
	 \end{cases}
\label{eq:igo_utility_tie}
\end{equation}
On the continuous domain, we can often assume that the Lebesgue measure of the level set of the objective function is $0$. Then, the utility function reads
\begin{equation}
W^f_{\theta^{(t)}}(x)=w(q_{\theta^{(t)}}^{\leq}(x)).
\label{eq:igo_utility}
\end{equation}

To update the parameter, the IGO takes the so-called \emph{natural gradient} ascent of $J_{\theta^{(t)}}$. The natural gradient of $J_{\theta^{(t)}}$ at $\theta$, denoted by $\tilde{\nabla} J_{\theta^{(t)}} (\theta)$, is given by the product of the inverse of the Fisher information matrix $F(\theta)$ at $\theta$ and the vanilla gradient of $J_{\theta^{(t)}}$ at $\theta$, denoted by $\nabla J_{\theta^{(t)}} (\theta)$, i.e., the vector of the partial derivative w.r.t.~each parameter. Given the natural gradient $\tilde{\nabla} l(\theta; x) = F^{-1}(\theta ) \nabla l(\theta; x)$ of the log-likelihood $l(\theta; x) = \ln p_\theta(x)$ at $\theta$, we have
\begin{equation}
\label{eq:natural_gradient}
\begin{split}
\tilde{\nabla} J_{\theta^{(t)}} (\theta) 
&= \tilde{\nabla} \int W^{f}_{\theta^{(t)}}(x) p_{\theta}(x) \rmd x 
= \int W^{f}_{\theta^{(t)}}(x) \tilde{\nabla}  p_{\theta}(x) \rmd x
\\
&= \int W^{f}_{\theta^{(t)}}(x) \big(\tilde{\nabla} l(\theta; x)\big) p_{\theta}(x) \rmd x \enspace.
\end{split}
\end{equation}
That is, the natural gradient of $\tilde{\nabla} J_{\theta^{(t)}}$ at $\theta$ is the expectation of the product of the utility $W^{f}_{\theta^{(t)}}(x)$ and the natural gradient of the log-likelihood $\tilde{\nabla} l(\theta; x)$ over $x \sim P_\theta$. 

In practice, the integral in \eqref{eq:natural_gradient} cannot be computed analytically since the objective function is black-box and we do not have full access on $f$. To instantiate the IGO algorithm, the natural gradient \eqref{eq:natural_gradient} at $\theta = \theta^{(t)}$ needs to be approximated. The natural gradient of $J_{\theta^{(t)}}$ at $\theta^{(t)}$ is approximated by Monte-Carlo using the candidate solutions $x_1,\dots,x_\lambda$ from the current distribution $P_{\theta^{(t)}}$ as
\begin{equation}
\tilde{\nabla} J_{\theta^{(t)}} (\theta^{(t)}) \approx \frac{1}{\lambda} \sum_{i=1}^\lambda W^{f}_{\theta^{(t)}}(x_i) \tilde{\nabla} l(\theta^{(t)}; x_i) \enspace.
\label{eq:ng_mc}
\end{equation}
Note that the natural gradient of the log-likelihood is independent of $f$ and is often computed explicitly, as we demonstrate two example cases later. On the other hand, we do not have full access to $W^{f}_{\theta^{(t)}}(x_i)$ and \del{we}{} need to approximate it by Monte-Carlo. 

To approximate the utility values for each $x_i$, the quantiles \eqref{eq:quantile} at each $x_i$ are approximated by the ranking of $x_i$ among $\{x_k\}_{k=1}^{\lambda}$, 
\begin{equation}
\label{eq:ranking}
\begin{split}
q_{\theta^{(t)}}^{\leq}(x_i) &\approx \frac1\lambda \rk^{\leq}(x_i) := \frac1\lambda \sum_{k=1}^{\lambda} \mathbb{I}\{{f(x_k) \leq f(x_i)}\} \enspace, \\
q_{\theta^{(t)}}^{<}(x_i) &\approx \frac1\lambda \rk^{<}(x_i) := \frac1\lambda \sum_{k=1}^{\lambda} \mathbb{I}\{f(x_k) < f(x_i)\}  \enspace,
\end{split}
\end{equation}
where $\rk^{\leq}(x_i)$ and $\rk^{<}(x_i)$ co\nnew{u}nt the numbers of weakly and strictly better candidate solutions, respectively, \nnew{and $\mathbb{I} \{\cdot\}$ denotes the indicator function}.
Let $W$ be the indefinite integral of $w$, i.e., $W(a) - W(b) = \int_{b}^{a} w(t) \rmd t$ for any $0 \leq b \leq a \leq 1$. Using the above approximation in \eqref{eq:igo_utility_tie}, we obtain an approximated utility 
\begin{equation}
W^f_{\theta^{(t)}}(x_i) \approx \hat{w}_i := \frac{W(\rk^{\leq}(x_i) / \lambda) - W(\rk^{<}(x_i) / \lambda)}{\rk^{\leq}(x_i)/\lambda - \rk^{<}(x_i) / \lambda} \enspace.
\label{eq:what}
\end{equation}
Note that if we assume that the probability of sampling the same function value is zero, $\hat{w}_i$ in \eqref{eq:what} takes only $\lambda$ values, i.e., $\hat{w}_i \in \{W(k/\lambda) - W((k-1)/\lambda) \mid k = 1,\dots, \lambda\}$.\footnote{Remark also that if we approximate the utility value with \eqref{eq:igo_utility} using the ranking \eqref{eq:ranking}, the approximation reads $w(\rk^{\leq}(x_i) / \lambda)$ or $w((\rk^{\leq}(x_i) - 1/2) / \lambda)$. The original IGO algorithm is defined with the latter utility approximation. The values are different from \eqref{eq:what}, but both cases take only $\lambda$ values, and one can take $w$ so that the weight values are equivalent.}


With the approximations \eqref{eq:ng_mc} and \eqref{eq:what}, we have an approximation of the natural gradient \eqref{eq:natural_gradient} at $\theta^{(t)}$ as $\tilde{\nabla} J_{\theta^{(t)}} (\theta^{(t)}) \approx \lambda^{-1} \sum_{i=1}^\lambda \hat{w}_i \tilde{\nabla} l(\theta^{(t)}; x_i)$. Then, we obtain the parameter update
\begin{equation}
\theta^{(t+1)} = \theta^{(t)} + \eta \sum_{i=1}^{\lambda} \frac{\hat{w}_i}{\lambda} \tilde{\nabla} l(\theta^{(t)}; x_i)
\enspace, 
\label{eq:igo-update}
\end{equation}
where $\eta$ denotes the learning rate for the parameter update.


\subsection{PBIL and compact GA}
\label{sec:pbil}
Population-based incremental learning (PBIL) is an example algorithm that is known to derive from the IGO framework with the family of Bernoulli distributions on $\bbX = \{ 0, 1 \}^d$ parameterized by the probability parameter $\theta \in \Theta = (0, 1)^d$, namely,
\begin{align}
p_{\theta} (x) = \prod_{i=1}^d \theta_i^{x_i} (1 - \theta_i)^{1 - x_i} \enspace,
\end{align}
where $\theta_i$, the $i$-th coordinate of $\theta$, represents the probability of \nnew{the} $i$-th bit of the sample $x$ being $1$. It is shown by \citet{Ollivier2017} that the natural gradient of the log-likelihood is $\tilde{\nabla} l(\theta; x) = (x - \theta)$. Then, \eqref{eq:igo-update} reads
\begin{align}
\theta^{(t+1)} = \theta^{(t)} + \eta \sum_{i=1}^{\lambda} \frac{\hat{w}_i}{\lambda} (x_i^{(t)} - \theta^{(t)})
\enspace.
\label{eq:pbil_update}
\end{align}
This update rule is equivalent to the one used in the PBIL. Moreover, considering the population size of $\lambda = 2$ and the weight function \nnew{is chosen} such that the better point receives $0.5$ and the worse point receives $-0.5$, \nnew{equation \eqref{eq:pbil_update}} recovers the compact GA (cGA).

\subsection{CMA-ES}
\label{sec:cma-es}

\subsubsection{The Pure Rank-$\mu$ Update CMA-ES}
The pure rank-$\mu$ update CMA-ES is considered an instantiation of the IGO algorithm. Given the multivariate Gaussian distribution $\mathcal{N}(\mm, \CC)$ parameterized by $\theta = (\mm,\CC)$, the natural gradients of log-likelihood for $\mm$ and $\CC$ are given by
\begin{align*}
& \tilde{\nabla}_{\mm} l(\theta; x) = x - \mm \\
& \tilde{\nabla}_{\CC} l(\theta; x) = (x - \mm) (x - \mm)^{\T} - \CC  .
\end{align*}
Introducing the different learning rates for the mean vector $\mm$ and the covariance matrix $\CC$, we get the following natural gradient update rules:
\begin{equation}
    \mm^{(t+1)} = \mm^{(t)} + \cm \sum_{i=1}^\lambda \frac{\hat{w}_i}{\lambda} (x_i^{(t)} - \mm^{(t)}) , 
    \label{eq:igo-m-update}
\end{equation}
\begin{equation}
	\CC^{(t+1)} = \CC^{(t)} + \cmu \sum_{i=1}^{\lambda} \frac{\hat{w}_i}{\lambda} \left( (x_i^{(t)} - \mm^{(t)}) (x_i^{(t)} - \mm^{(t)} )^{\T} - \CC^{(t)} \right) ,
	\label{eq:igo-c-update}
\end{equation}
where $\cm$ and $\cmu$ are the learning rates for $\mm$ and $\CC$, respectively, and usually $\cm = 1$.

In the standard CMA-ES, the weights are defined as follows. Let $\rk(x_i^{(t)})$ denote the ranking of the candidate solution $x_i^{(t)}$ among $\lambda$ candidate solutions. Each weight defined in \new{\citep{Hansen2014}} is
\begin{equation}
    \frac{\hat{w}_{i}}{\lambda} = \frac{\max(0, \ln (\frac{\lambda+1}{2}) - \ln (\rk(x_i^{(t)})))}{\sum_{j=1}^{\lambda} \max(0, \ln (\frac{\lambda+1}{2}) - \ln (j))} \enspace.
   \label{eq:cma-weight}
\end{equation}

\subsubsection{Rank-one Update}
The update rule of the covariance matrix $\CC$ consists of two components: \nnew{the} rank-one update and \nnew{the} rank-$\mu$ update.
The rank-one update accelerates the update of the covariance matrix using the so-called evolution path which is the cumulation of the consecutive steps.
The evolution path for the rank-one update, $\pc$, is updated with the following formula
\begin{equation}
    \pc^{(t+1)} = (1 - \cc) \pc^{(t)} + \sqrt{\cc (2 - \cc)\mueff} \sum_{i=1}^{\lambda}\frac{\hat{w}_i}{\lambda}(x_i^{(t)} - \mm^{(t)}),
    \label{eq:evolution-path}
\end{equation}
where $\cc$ is the cumulation parameter for the evolution path and $\mueff = \nnew{\lambda^2 (\sum_{i=1}^{\lambda} \hat{w}_i^2)^{-1}}$ is the so-called effective variance selection mass.
\yohe{Comment: $\mu_w = \frac{(\sum\lvert w_i \rvert)^2}{\sum\lvert w_i \rvert^2}$ may be the actual definition when $\sum\lvert w_i \rvert \neq 1$. But it doesn't matter here.}%
The update rule of the covariance matrix $\CC$ with the rank-one and rank-$\mu$ update is then
\begin{align}
		\CC^{(t+1)} = \CC^{(t)}
		+ & \underbrace{\cone (\pc^{(t+1)} (\pc^{(t+1)})^\T - \CC^{(t)})}_{\text{rank-one update}} \notag \\
		& + \underbrace{\cmu \sum_{i=1}^{\lambda} \frac{\hat{w}_{i}}{\lambda} \left( (x_i^{(t)} - \mm^{(t)}) (x_i^{(t)} - \mm^{(t)} )^{\T} - \CC^{(t)} \right)}_{\text{rank-$\mu$ update}} ,
	\label{eq:hybrid}
\end{align}
where $\cone$ and $\cmu$ are the learning rates of the rank-one and rank-$\mu$ updates for $\CC$, respectively.
The rank-one update enlarges the eigenvalue of $\CC$ correspond\new{ing} to the direction of the evolution path, and the rank-$\mu$ update is considered the natural gradient ascent of the expectation of the transformed objective function as described in Section \ref{sec:igo}.
The CMA-ES described above with $\cone = 0$ is called the pure rank-$\mu$ update CMA-ES.


\subsection{Importance Mixing}
An idea of reusing previously generated solutions has been introduced in \citep{Yi2009a,Yi2009}, called importance mixing. It creates the set of points that can be considered as taken from the current distribution $\pthetat$. Provided a population of $\lambda$ points drawn from $p_{\theta^{(t-1)}}$, it accepts each point $x$ in the population with probability $\min\{1, (1 - \alpha) \pthetat(x) / p_{\theta^{(t-1)}}(x)\}$ into the new population. Let $\lambda' \leq \lambda$ be the number of accepted points, \nnew{then the method samples} a point $x$ from $\pthetat$ and accepts it with probability $\max\{\alpha, 1 - p_{\theta^{(t-1)}}(x)/\pthetat(x)\}$ into the new population\nnew{; this procedure is repeated} until $\lambda - \lambda'$ points are accepted. Then, the new population is considered as being distributed as $\pthetat$. Since \nnew{the importance mixing was} introduced to reduce the number of points for which the function value is evaluated at each iteration, the estimation variance of the natural gradient does not necessarily lessen. In Section~\ref{sec:exp_gaussian_reuse_igo}, we \nnew{will} compare our proposed strategies with the importance mixing method. The parameter $\alpha$, called the minimal refresh rate, is set to $\alpha = 0.0$ to reduce the function evaluations as much as possible in the experiments.\footnote{The minimal refresh rate $\alpha=0.01$ is recommended in \citep{Yi2009a} for the \emph{efficient natural evolution strategy} (eNES). In our preliminary experiments, we did not observe \nnew{a} statistically significant difference between $\alpha = 0.01$ and $\alpha = 0.0$, except that the former exhibits higher variance in the success probability on multimodal functions.} 

\section{Sample Reuse in the IGO}
\label{sec:reuse-igo}

In this section, we propose a sample reuse \del{mechanism}{}\nnew{technique} for the IGO framework based on importance sampling. Importance sampling is a technique to estimate the expectation over a probability distribution by using samples drawn from a different probability distribution. We mix the current and past samples as \nnew{if they are} drawn from the mixture of the current and past distributions and estimate the natural gradient by Monte-Carlo using the samples from the mixture. 

\subsection{Importance Sampling} \label{sec:importance-sampling}
Suppose that we have $K + 1$ probability distributions $p^k$ ($k = 0, 1, \dots,K$) and $\lambda$ independent samples $x^k_i$ ($i=1,\dots,\lambda$) \nnew{drawn} from each probability distribution $p^k$. Our objective here is to estimate the expectation $\int g(x) p^{0}(x) \rmd x$ by using \nnew{the} $\lambda (K+1)$ samples $x^k_i$. In the IGO, $p^0$ and $p^k$ corresponds to the current distribution and the distribution of previous $k$-th iteration. The simplest way is to apply the Monte-Carlo estimate by using only the samples $x^0_i$ from $p^{0}$: 
\begin{equation}
\frac{1}{\lambda}\sum_{i=1}^{\lambda} g(x^0_i) \enspace,
\label{eq:monte-carlo}
\end{equation}
implying that we discard all the other samples $x^k_i$ ($k \geq 1$), despite that they may help to estimate the quantity more accurately. To utilize all the possible samples, we employ the idea of importance sampling.

Let $\bar{p}$ be the mixture $\bar{p}(x) = (K+1)^{-1}\sum_{k=0}^{K} p^k(x)$. One can rewrite the expectation as
\begin{equation}
\int g(x) p^{0}(x) \rmd x = \int g(x) \frac{p^{0}(x)}{\bar{p}(x)} \bar{p}(x) \rmd x \enspace.
\label{eq:is-mixture-exp}
\end{equation}
Considering the set of points $x^k_i$ to be sampled from $\bar{p}$, we can approximate the RHS of the above equality by the average
\begin{equation}
\frac{1}{\lambda(K+1)}\sum_{k=0}^{K}\sum_{i=1}^{\lambda}g(x^k_i) \frac{p^{0}(x^k_i)}{\bar{p}(x^k_i)}
\enspace,
\label{eq:is-mixture}
\end{equation}
which is an unbiased estimator of \eqref{eq:is-mixture-exp} \nnew{as} shown in \cite[Section 3.2]{Veach1995}, i.e., the expected value of \eqref{eq:is-mixture} equals to \eqref{eq:is-mixture-exp}. In this way, we can utilize all the samples without introducing any bias and expect that this leads to more accurate estimation than the simple Monte-Carlo using only $x^0_i$. This is the key idea of our work. Note that the case $K = 0$ recovers \eqref{eq:monte-carlo}.%

One can create a different unbiased estimator that uses all the possible samples. Instead of considering the mixture $\bar{p}$, one computes $K+1$ unbiased estimators $\lambda^{-1}\sum_{i=1}^{\lambda}g(x^k_i) p^0(x^k_i) / p^k(x^k_i)$ and averages them, resulting in
\begin{equation}
  \frac{1}{\lambda(K+1)}\sum_{k=0}^{K}\sum_{i=1}^{\lambda}g(x^k_i) \frac{p_0(x^k_i)}{p_k(x^k_i)} \enspace.
  \label{eq:k_monte-carlo}
\end{equation}
This is also an unbiased estimator of \eqref{eq:is-mixture-exp}. We can prove that \eqref{eq:is-mixture} has a smaller estimation variance than \eqref{eq:k_monte-carlo}, which is also claimed by \cite{Shelton2001a,Shelton2001}. See Appendix~\ref{apdx:proof} for the proof.

\subsection{Sample Reuse in the IGO}

We introduce \eqref{eq:is-mixture} into the IGO framework to estimate the natural gradient $\tilde{\nabla} J_{\theta^{(t)}} (\theta^{(t)})$ using the samples $x^{(t-k)}_{i}$ ($i = 1, \dots, \lambda$) from the current and past $K$ distributions, $p^{t - k}$ ($k = 0, 1, \dots, K$). Let $\bar{p}(x) = (K+1)^{-1}\sum_{k=0}^{K} p^{(t-k)}(x)$ be the mixture of the current distribution and $K$ previous distributions.

We first estimate the utility $W^f_{\theta^{(t)}}(x_i^{(t-k)})$. Applying the formula \eqref{eq:is-mixture} \nnew{to \eqref{eq:ranking}}, we obtain the following approximation of the quantiles \eqref{eq:quantile}:
\begin{equation}
\begin{split}
  \bar{q}^{\leq} (x^{(t-k)}_i) := \sum_{l=0}^{K}\sum_{j=1}^{\lambda} \frac{\ind{ f(x^{t-l}_j) \leqslant f(x^{(t-k)}_i)}}{\lambda(K+1)} \frac{p^{(t)}(x^{t-l}_j)}{\bar{p}(x^{t-l}_j)}
  \enspace,
\\
\bar{q}^{<} (x^{(t-k)}_i) := \sum_{l=0}^{K}\sum_{j=1}^{\lambda} \frac{\ind{ f(x^{t-l}_j) < f(x^{(t-k)}_i)}}{\lambda(K+1)} \frac{p^{(t)}(x^{t-l}_j)}{\bar{p}(x^{t-l}_j)}
\enspace.
\end{split}
\label{eq:q-is}
\end{equation}
The utility value $W^f_{\theta^{(t)}}(x_i^{(t-k)})$ assigned to each candidate solution is then approximated in the same manner as in \eqref{eq:what}, namely,
\begin{equation}
W^f_{\theta^{(t)}}(x_i^{(t-k)}) \approx \hat{w}_i^{(t-k)} := \frac{W(\bar{q}^{\leq} (x^{(t-k)}_i)) - W(\bar{q}^{<} (x^{(t-k)}_i))}{\bar{q}^{\leq} (x^{(t-k)}_i) - \bar{q}^{<} (x^{(t-k)}_i)} \enspace.
\label{eq:weight_estimated}
\end{equation}
\nnew{Unlike} the original estimates \eqref{eq:ranking} of the quantiles, \eqref{eq:q-is} is not guaranteed to live in $[0, 1]$. It is guaranteed to be positive, but can be greater than $1$. Moreover, \eqref{eq:q-is} can take an arbitrary positive real value, whereas \eqref{eq:ranking} takes only $\lambda$ different values. Therefore, we need to prepare (input) a function $W: [0, \infty) \to \R$ rather than $\lambda$ values $W(k/\lambda) - W((k-1)/\lambda)$.



Next we estimate the natural gradient \eqref{eq:natural_gradient} using \eqref{eq:weight_estimated}. The natural gradient \eqref{eq:natural_gradient} at $\theta^{(t)}$ is of the form \eqref{eq:is-mixture-exp} with $g(x) = W^{f}_{\theta^{(t)}}(x) \tilde{\nabla} l(\theta^{(t)}; x)$. With an approximation of $W^{f}_{\theta^{(t)}}(x_i^{(t-k)})$ for each $i$ and $k$ defined in \eqref{eq:weight_estimated}, we can apply the formula \eqref{eq:is-mixture} and obtain the estimate
\begin{equation}
\begin{split}
  \tilde{\nabla} J_{\theta^{(t)}} (\theta^{(t)}) = \frac{1}{\lambda (K+1)} \sum_{k=0}^{K} \sum_{i=1}^{\lambda} \hat{r}(x_i^{(t-k)}) \tilde {\nabla} l(\theta^{(t)}; x_i^{(t-k)}) \enspace,
\\
  \quad \text{where} \quad
  \hat{r}(x_i^{(t - k)}) = \hat{w}_i^{(t - k)} \frac{p_{\theta^{(t)}}(x_i^{(t - k)})}{\bar{p}(x_i^{(t - k)})}
  \enspace.
\end{split}
\label{eq:is-ng}
\end{equation}

\del{We remark that the proposed estimator \eqref{eq:is-ng} as well as the original estimator (the second term on the RHS of \eqref{eq:igo-algo}) is biased; its expectation differs from \eqref{eq:natural_gradient} since the same samples are used to estimate $W^f_{\theta^{(t)}}$ and $\tilde\nabla J_{\theta^{(t)}}$, but they are consistent; the bias vanishes and both estimates converge to \eqref{eq:natural_gradient} w.p.1 as $\lambda \to \infty$. See Remark~2 of \cite{Akimoto2012ppsn} for the impact of the bias.}{}

\paragraph{Remark}
The sum of the weights \eqref{eq:what} is always $\lambda(W(1) - W(0))$. A similar property holds for the weights multiplied by the likelihood ratio $\hat{r}(x_i^{(t - k)})$ in \eqref{eq:is-ng}. Let $M$ be the number of distinct values in $(\bar{q}^{\leq} (x^{(t-k)}_i))_{i=1,\dots,\lambda}^{k=1,\dots,K}$ and let $\bar{q}^{\leq}_{i:M}$ and $\bar{q}^{<}_{i:M}$ be the $i$-th smallest values among the $M$ distinct values of $(\bar{q}^{\leq} (x^{(t-k)}_i))_{i=1,\dots,\lambda}^{k=1,\dots,K}$ and $(\bar{q}^{<} (x^{(t-k)}_i))_{i=1,\dots,\lambda}^{k=1,\dots,K}$, respectively. Then, from \eqref{eq:q-is} we have $\bar{q}^{\leq}_{i:M} = \bar{q}^{<}_{i+1:M}$ for $i = 1,\dots,M-1$ and $\bar{q}^{<}_{1:M} = 0$. Moreover, let $m_i$ be the number of candidate solutions such that $\bar{q}^{\leq} (x^{(t-k)}_j) = \bar{q}^{\leq}_{i:M}$ and let $y_{i,j}$ denote such candidate solutions (for $j = 1,\dots, m_i$). We then have
\begin{align*}
  \MoveEqLeft[2]\frac{1}{\lambda (K+1)} \sum_{k=0}^{K} \sum_{i=1}^{\lambda} \hat{r}(x_i^{(t - k)})
  \\
  &= \frac{1}{\lambda (K+1)} \sum_{k=0}^{K} \sum_{i=1}^{\lambda} \hat{w}_i^{(t - k)} \frac{p_{\theta^{(t)}}(x_i^{(t - k)})}{\bar{p}(x_i^{(t - k)})}
  \\
  &= \frac{1}{\lambda (K+1)} \sum_{k=0}^{K} \sum_{i=1}^{\lambda} \frac{W(\bar{q}^{\leq} (x^{(t-k)}_i)) - W(\bar{q}^{<} (x^{(t-k)}_i))}{\bar{q}^{\leq} (x^{(t-k)}_i) - \bar{q}^{<} (x^{(t-k)}_i)} \frac{p_{\theta^{(t)}}(x_i^{(t - k)})}{\bar{p}(x_i^{(t - k)})}    
  \\
  &= \frac{1}{\lambda (K+1)} \sum_{i=1}^{M} \sum_{j=1}^{m_i} \frac{W(\bar{q}^{\leq}_{i:M}) - W(\bar{q}^{<}_{i:M})}{\frac{m_i}{\lambda (K+1)}  \frac{p_{\theta^{(t)}}(y_{i,j})}{\bar{p}(y_{i,j})}} \frac{p_{\theta^{(t)}}(y_{i,j})}{\bar{p}(y_{i,j})}    
  \\
  &= \sum_{i=1}^{M} W(\bar{q}^{\leq}_{i:M}) - W(\bar{q}^{<}_{i:M})
  \\
  &= W(\bar{q}^{\leq}_{M:M}) - W(\bar{q}^{<}_{1:M}) =  W(\bar{q}^{\leq}_{M:M}) - W(0)
  \enspace.
\end{align*}
Since $\bar{q}^{\leq}_{M:M}$ is not guaranteed to be $1$, we may have \nnew{a} changing sum of the weights, which may affect the learning rate. However, as long as $w(x) = 0$ for $x > 1/2$ such as the weights standard for the CMA-ES, we will almost always have a constant sum of weights since the probability of $\bar{q}^{\leq}_{M:M}$ being smaller than $1/2$ is quite small. \yohe{the same holds for the other type of IS.}

\section{Application to the PBIL and the CMA-ES}
\label{sec:app}

Now that the importance sampling technique has been introduced in the IGO framework, its application to the PBIL and the pure rank-$\mu$ update CMA-ES is straightforward. Here we introduce the importance sampling in the PBIL and the CMA-ES combined with the rank-one update.

\subsection{Sample Reuse in the PBIL}
\label{sec:reuse-pbil}

We introduce the importance sampling to the PBIL (\nnew{including} compact GA). Let us consider the following step function as the utility function:
\begin{equation}
w(s) =
	\begin{cases}
	\frac{1}{2T} & (s \leq T) \\
	 0 & (T < s \leq 1 - T) \\
	 -\frac{1}{2T} &  (1 - T < s) \enspace,
	 \end{cases}
\end{equation}
where $T$ represents the threshold parameter, and $w(s)$ satisfies $\int_0^1 \lvert w(s) \rvert \rmd s = 1$ and  $\int_0^1  w(s) \rmd s = 0$. Then, one can define
\begin{equation}
  \label{eq:pbilw}
  W(s) = \int_{0}^{s} w(t) \rmd t =
  \begin{cases}
    \frac{s}{2T} & (\new{s} \leq T) \\
    \frac{T}{2T} & (T < s \leq 1 - T) \\
    \frac{1 - s}{2T} &  (1 - T < s) \enspace.
  \end{cases}
\end{equation}
The utility value $W^f_{\theta^{(t)}}(x_i^{(t-k)})$ is approximated by \nnew{the} importance sampling scheme \eqref{eq:weight_estimated}.

In the experiment, we set $T = 0.25$. When $\lambda=2$ and $K=0$ (i.e.,\ importance sampling is not used) and the utility value is approximated by Monte-Carlo using only current samples, the weight values become $\hat{w}_{1:\lambda}^{(t)} = 1$ and $\hat{w}_{2:\lambda}^{(t)} = - 1$ if $f(x_{1})$ and $f(x_{2})$ are distinct and $\hat{w}_{1}^{(t)} = \hat{w}_{2}^{(t)} = 0$ for the tie case. In this setting, the algorithm is exactly the same as the compact GA.


Introducing the natural gradient of the log-likelihood of the Bernoulli distribution into \eqref{eq:is-ng}, we get the importance sampling version of the parameter update rule as follows:
\begin{align}
\theta^{(t+1)} = \theta^{(t)} + \eta \sum_{k=0}^{K} \sum_{i=1}^{\lambda} \frac{\hat{r}(x_i^{(t-k)})}{\lambda (K+1)} (x_i^{(t-k)} - \theta^{(t)})\enspace,
\end{align}
where $\hat{r}(x_i^{(t-k)})$ is the product of $\hat{w}_i^{(t - k)}$ and the likelihood ratio, defined in \eqref{eq:is-ng}.

\subsection{Sample Reuse in the Rank-$\mu$ Update CMA-ES}
\label{sec:reuse-cma}
We introduce the importance sampling to the CMA-ES as described in Section~\ref{sec:cma-es}. More precisely, we replace the mean vector update \eqref{eq:igo-m-update} and the rank-$\mu$ update of the covariance matrix in \eqref{eq:hybrid} with novel ones using the samples from the current and past $K$ iteration.

As discussed in Section~\ref{sec:igo}, the pure rank-$\mu$ update CMA-ES is considered as an instantiation of the IGO algorithm with the multivariate Gaussian distribution. Therefore, we can simply replace the second term on the RHS of \eqref{eq:igo-m-update} and the rank-$\mu$ update of the covariance matrix in \eqref{eq:hybrid} with the novel natural gradient estimate \eqref{eq:is-ng}. All we need is to choose the function $w$. We choose $w(s) = -2 \ln (2 s) \ind{s \leq 1/2}$, since $\hat{w}_i$ defined in \eqref{eq:cma-weight} is considered an approximation of $w((\rk(x_i^{(t)}) - 1/2)/\lambda)$ for large $\lambda$.\footnote{%
  The weight $\hat{w}_i$ defined in \eqref{eq:cma-weight} is rewritten as
\begin{align*}
  \hat{w}_i =   \frac{\max(0, \ln (\frac{\lambda+1}{2}) - \ln (\rk(i))}{ \frac{1}{\lambda} \sum_{j=1}^{\lambda} \max(0, \ln (\frac{\lambda+1}{2}) - \ln (j))} 
  = \frac{ \left( \ln(1+1/\lambda) - \ln(2\rk(x_i)/\lambda ) \right) \ind{\rk(x_i)/\lambda \leq 1/2}}{ \sum_{j=1}^{\lambda} \frac{1}{\lambda} \left( \ln(1+1/\lambda) - \ln(2j/\lambda ) \right) \ind{j/\lambda \leq 1/2} }\enspace.
\end{align*}
The denominator converges to $- \int_{0}^{1/2} \ln 2z \rmd z = 1/2$ as $\lambda \to \infty$, whereas the numerator is approximated by $w((\rk(x_i^{(t)}) - 1/2)/\lambda)$ and converges to $w(q_{\theta^{(t)}}^{\leq}(x_i^{(t)}))$ as $\lambda \to \infty$. 
}
Then, one can define
\begin{equation}
  \label{eq:cmaw}
  W(s) = \int_{0}^{s} w(t) \rmd t =
  \begin{cases}
    0 & (s = 0) \\
    \new{2}s - 2 s \ln(2s) & (s \leq 1/2) \\
    \new{1} &  (1/2 < s) \enspace.
  \end{cases}
\end{equation}
As in the PBIL case, the utility value $W^f_{\theta^{(t)}}(x_i^{(t-k)})$ is approximated by the importance sampling scheme \eqref{eq:weight_estimated}. The IGO update using the important sampling \eqref{eq:is-ng} reads
\begin{align}
 m^{(t+1)} &= m^{(t)} + \cm \sum_{k=0}^{K} \sum_{i=1}^{\lambda} \frac{\hat{r}(x_i^{(t-k)})}{\lambda (K+1)} ( x_i^{(t-k)} - m^{(t)}), \label{eq:ngm} \\ 
 C^{(t+1)} &= C^{(t)} + \cmu \sum_{k=0}^{K} \sum_{i=1}^{\lambda}  \frac{\hat{r}(x_i^{(t-k)})}{\lambda (K+1)} \left( (x_i^{(t-k)} - m^{(t)}) (x_i^{(t-k)} - m^{(t)} )^{\T} - C^{(t)} \right) \enspace,
\label{eq:ngc}
\end{align}
where $\hat{r}(x_i^{(t-k)})$ is as defined in \eqref{eq:is-ng}.

Note that if $K = 0$, \eqref{eq:ngm} and \eqref{eq:ngc} are equivalent to \eqref{eq:igo-m-update} and \eqref{eq:hybrid} with $\cone = 0$, except that $\hat{w}_i^{(t)} / \lambda$ differs from $w_i$ defined in \eqref{eq:cma-weight}. In the preliminary experiments, we have observed that the sum of \new{$\hat{r}(x_i^{(t-k)}) / \lambda (K+1)$ is almost always} one. \del{Its impact is observed in the next section.}{}

\del{
We combine the rank-one update to \eqref{eq:ngc}. We simply add the second term on the RHS of \eqref{eq:hybrid} to \eqref{eq:ngc}. To update the evolution path, we solely uses the current samples and use $w_{\rk(x_i^{(t)})}$ as the weight for each $x_i^{(t)}$ rather than $\hat{w}_{i}^{(t)}$. It means, the evolution path is updated as it is done in \eqref{eq:evolution-path}. Since the evolution path itself accumulates the past information, the rank-one update is considered utilizing the past samples. Therefore, the proposed algorithm exploits the past information in different ways to update the covariance matrix, and we expect the synergy.}

\paragraph{Implementation Remark}
To implement the proposed method, one needs to keep the mean vectors and the covariance matrices of the past $K$ distributions to compute the likelihood ratio $p_{\theta^{(t)}}(x_i^{(t-k)}) / \bar{p}(x_i^{(t-k)})$, which requires $O(K d^2)$ additional memory space. For efficient and numerically stable computation, we keep the log-likelihood of each point, $l_{i}^{k,l} = \ln p_{\theta^{(t-k)}}(x_{i}^{t-l})$, for each $i = 1,\dots,\lambda$ and $k, l = 0, \dots, K$, which requires $O(\lambda K^2)$ additional space. The likelihood ratio is then computed as 
$p_{\theta^{(t)}}(x_i^{(t-k)}) / \bar{p}(x_i^{(t-k)}) = (K+1)  ( \sum_{l=0}^{K} \exp (  l_i^{l,k} - l_{i}^{0,k} ) )^{-1}$. The computational complexity at each iteration is then $O(\lambda K d^2)$.



\yohe{population size one seems not working properly}


\providecommand{\onemax}{\textsc{OneMax}}
\providecommand{\leadingones}{\textsc{LeadingOnes}}
\section{Experimental Evaluation of the Sample Reuse PBIL}
\label{sec:exppbil}
To evaluate the effect of the sample reuse based on the importance sampling in the IGO framework, we conduct experiments in this and the next sections. 
In this section, we evaluate the proposed sample reuse IGO with the Bernoulli distribution.
As we described in Section \ref{sec:pbil}, the IGO with the Bernoulli distribution recovers PBIL and cGA when $\lambda = 2$.
Here, we focus on the cGA which is the simplest case and verify the effect of the proposed sample reuse \del{method}{}\del{\new{mechanism}}{}\nnew{technique}.

\subsection{Benchmark Functions and Experimental Setting}
We use two commonly used binary benchmark functions: \del{the one max}{}\new{\onemax{}} and \del{leading one}{}\new{\leadingones{}} functions. The $d$ bits \del{one max}{}\new{\onemax{}} function is defined by
\begin{equation}
f_{\mathrm{OneMax}}(x) = \sum_{i=1}^{d} x_i \enspace,
\end{equation}
and the $d$ bits \del{leading one}{}\new{\leadingones{}} function is defined by
\begin{equation}
f_{\mathrm{LeadingOnes}}(x) = \sum_{j=1}^{d} \prod_{i=1}^{j} x_i \enspace,
\end{equation}
where $x_i$ denotes the $i$-th bit \new{of a bit string $x$}. These \del{problems }{}are maximization problem\new{s}, and \del{the both optimum values are $d$}{}\new{both functions have the optimal value of $d$ at $x = (1, \dots, 1)$}. The \onemax{} is a separable problem whereas the \leadingones{} is a non-separable problem.

The initial parameters $\theta$ of the Bernoulli distribution are set to $0.5$. Each run \del{that finds the optimum value }{}is regarded as a success \new{if the optimal solution is sampled within $3d \times 10^2$ and $4d \times 10^4$ function evaluations on \onemax{} and \leadingones{}, respectively.}\del{whereas it is regarded as a failure after $3d \times 10^2$ and $4d \times 10^4$ function evaluations for one max and leading one problems, respectively.}{} We conduct $50$ independent runs for each setting. To keep the possibility to generate arbitrary bits, we restrict the parameters of the Bernoulli distribution within the range of $[1/d, 1-1/d]$. \del{That is, we set the boundary value to the parameter $\theta$ when it will be updated beyond the range. }{}\new{The learning rate and the number of iterations for sample reuse are set as}\del{We vary the learning rate as}{} $\eta \in \{ 1/d, 2/d, 4/d, 8/d, 16/d \}$ and \del{the number of iterations for sample reuse as }{}$K \in \{ 0, 1, 2, 3, 5, 7, 9 \}$. The algorithm with $K = 0$ \del{is exactly same with}{}\new{recovers} the \new{standard} cGA. We evaluate the search performance of each method by the average number of function evaluations over successful runs divided by the success probability.

\subsection{Results and Discussion}
\begin{figure}[tb]
\centering
\begin{subfigure}[]{0.49\linewidth}
  \centering  
  \includegraphics[height=4.7cm]{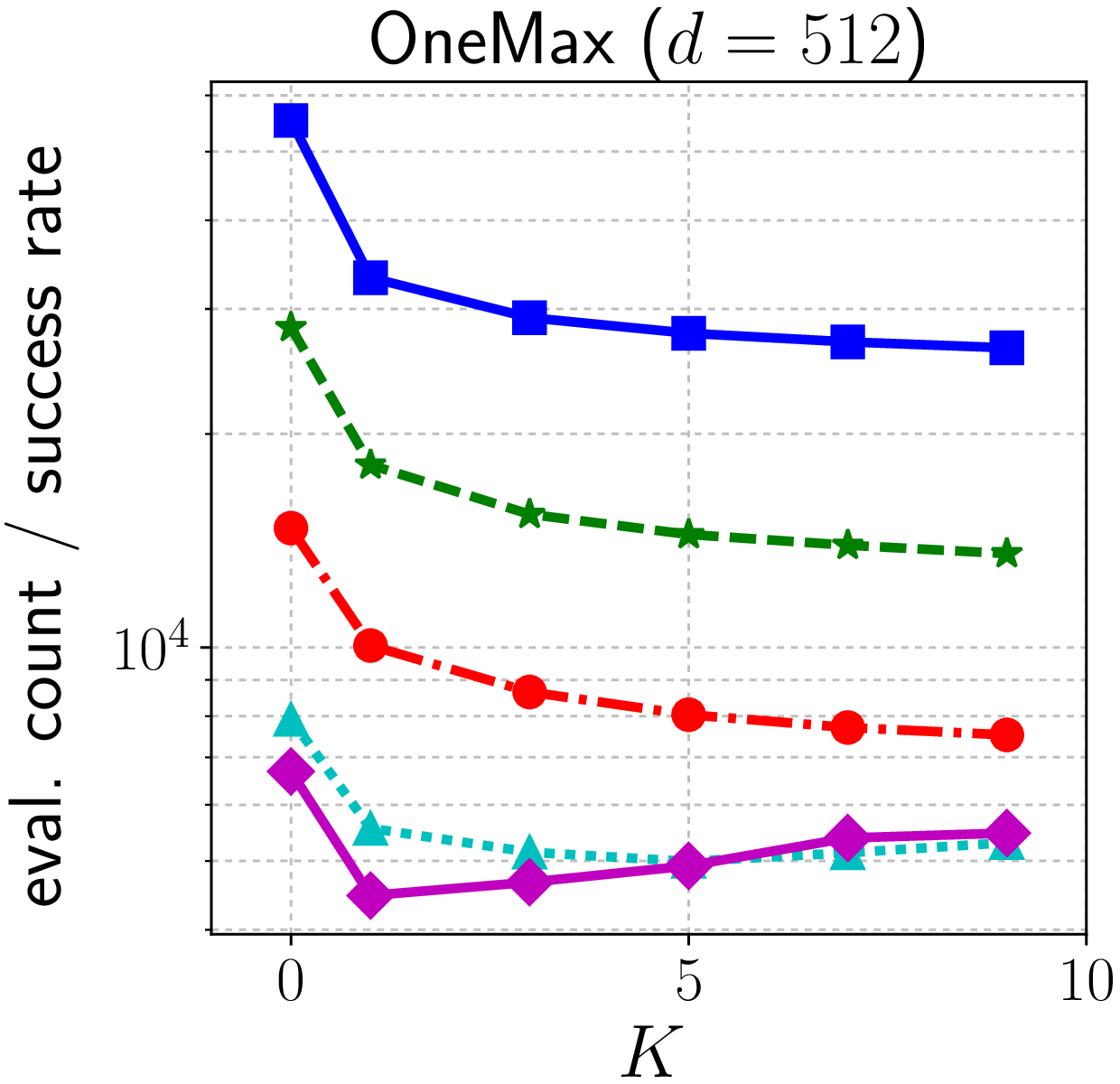}
\end{subfigure}
\begin{subfigure}[]{0.49\linewidth}
  \centering  
  \includegraphics[height=4.7cm]{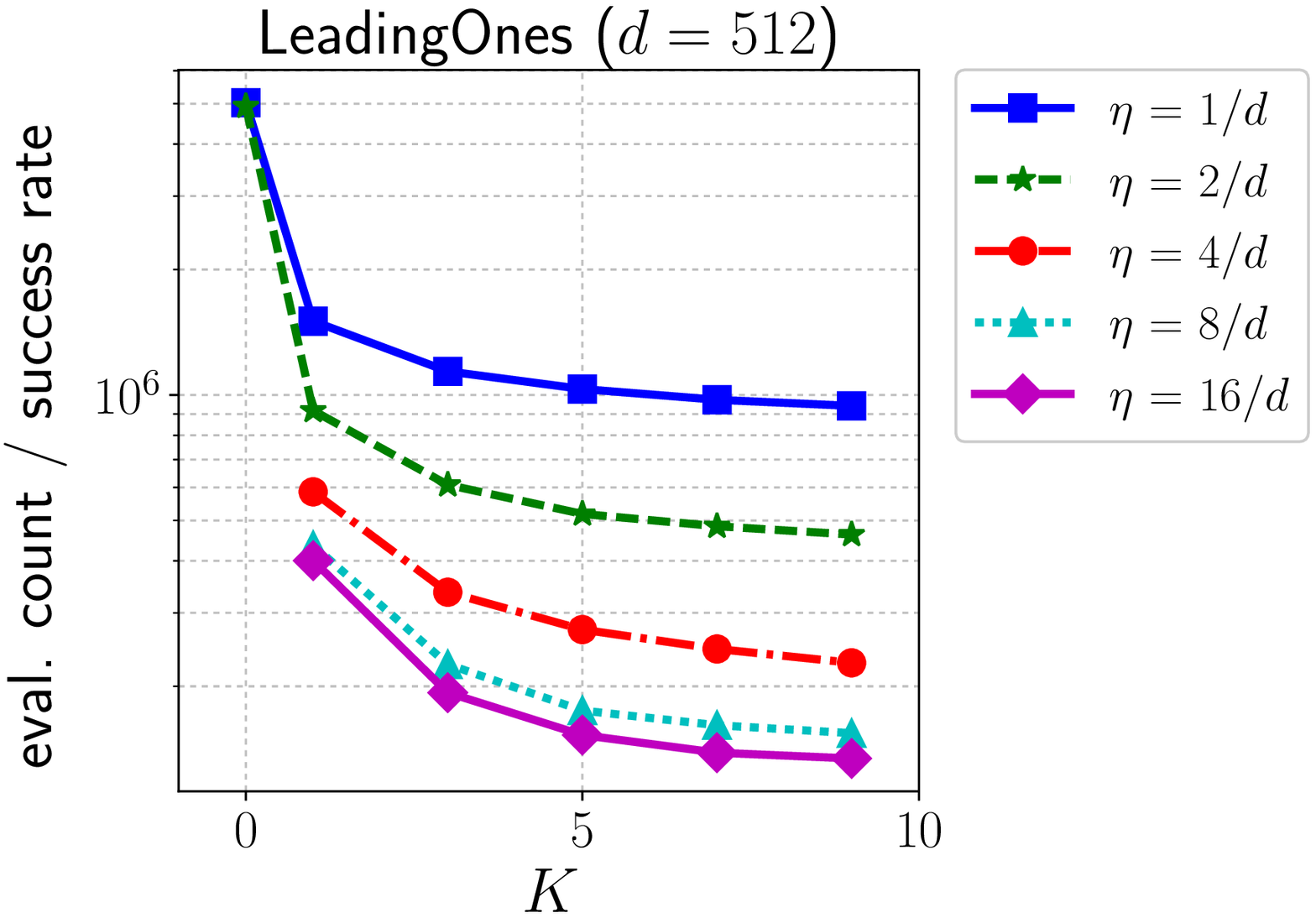}
\end{subfigure}
\caption{The search performance for various numbers of sample reuse iterations ($K \in \{0, 1, 2, 3, 5, 7, 9\}$). The results on the $512$-bit \onemax{} (left) and \leadingones{} (right) problems are displayed for the different learning rates ($\eta \in \{1/d, 2/d, 4/d, 8/d, 16/d\}$).}
\label{fig:bernoulli_k}
\end{figure}

\begin{figure}[tb]
\centering
\begin{subfigure}[]{0.49\linewidth}
  \centering  
  \includegraphics[height=4.7cm]{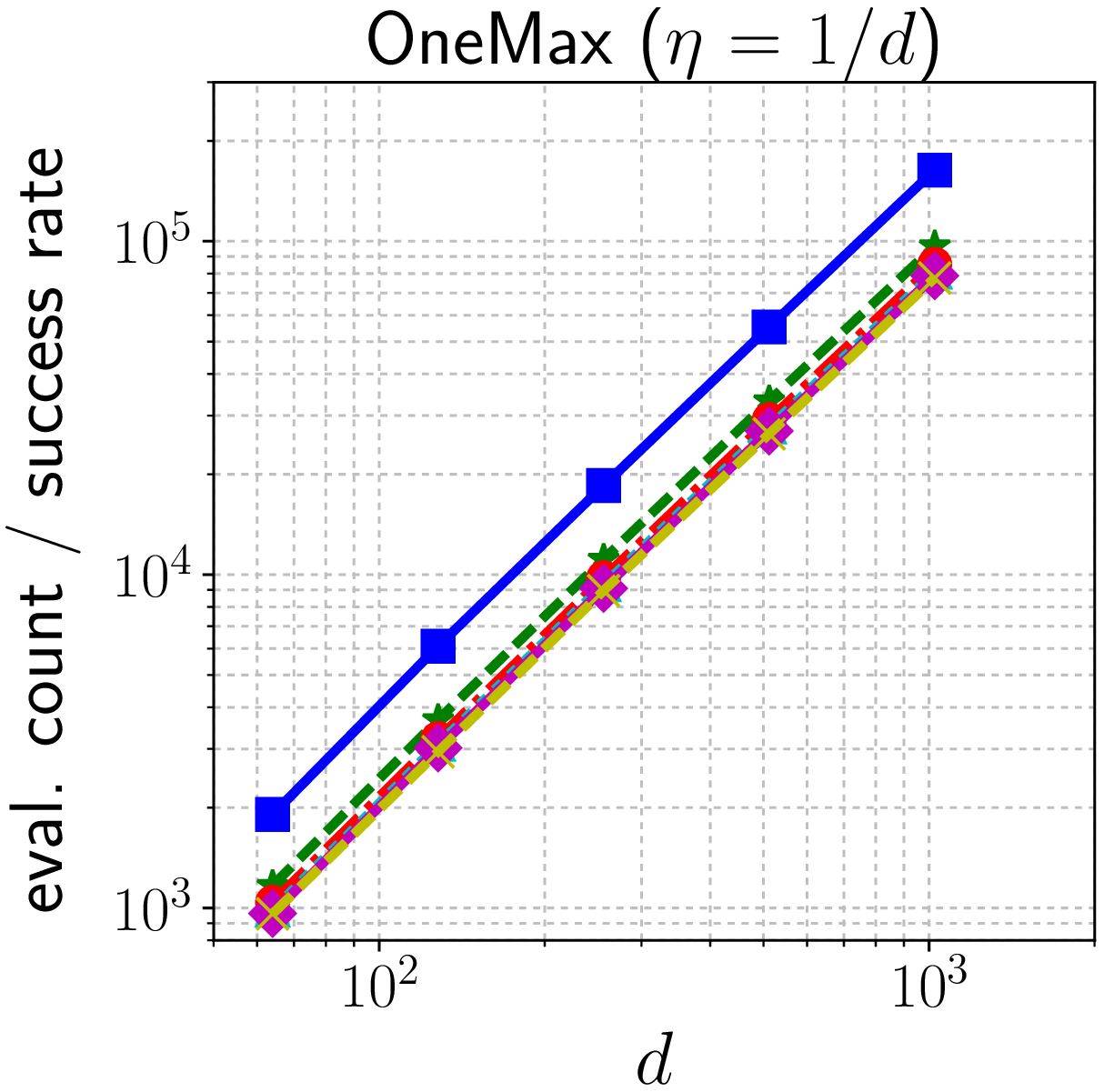}
\end{subfigure}
\begin{subfigure}[]{0.49\linewidth}
  \centering  
  \includegraphics[height=4.7cm]{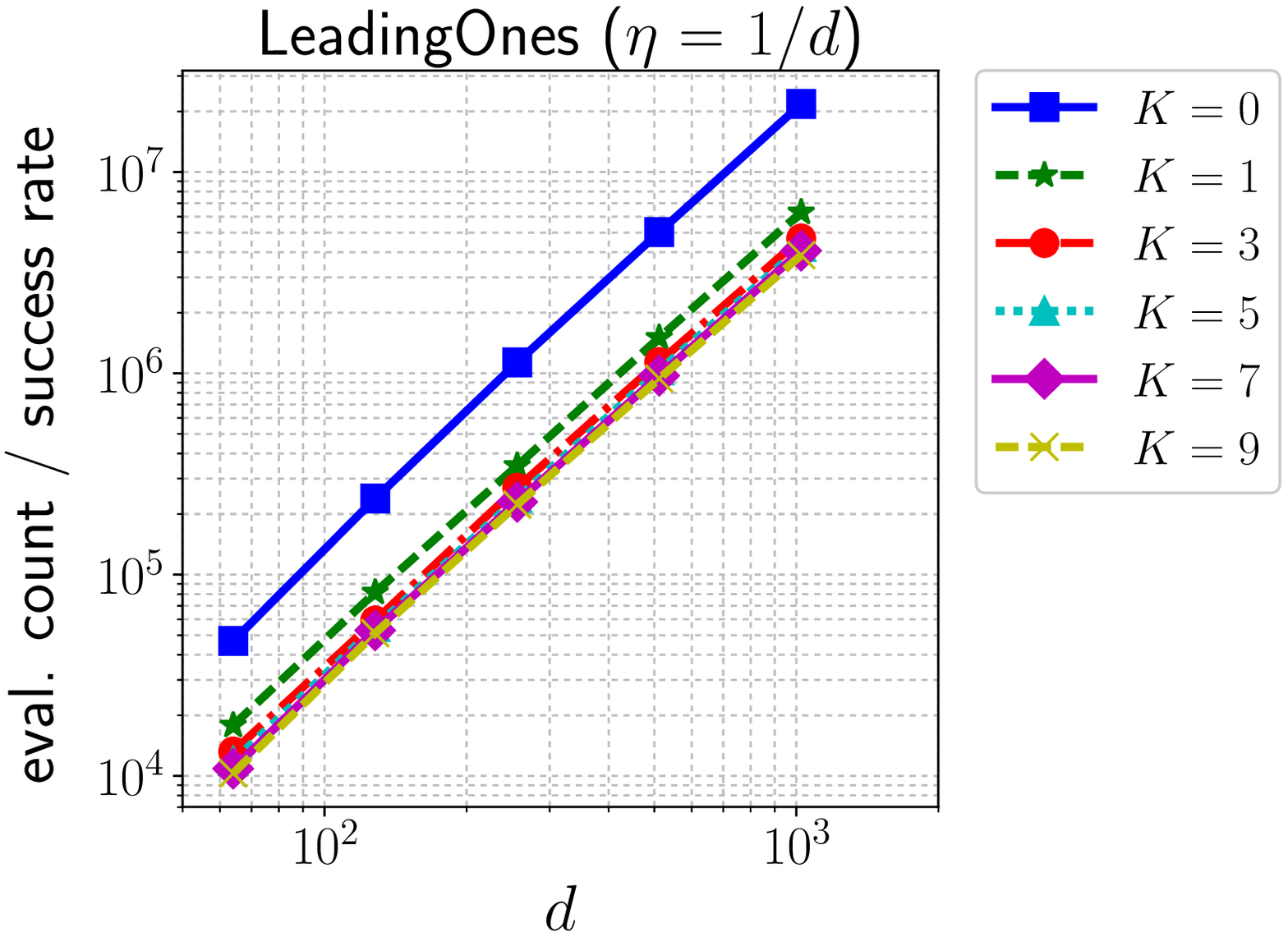}
\end{subfigure}
\caption{The search performance for various problem dimensions $d$. The results on the \onemax{} (left) and \leadingones{} (right) problems are displayed with the different numbers of sample reuse iterations ($K \in \{0, 1, 2, 3, 5, 7, 9\}$). The learning rate is set to $\eta = 1/d$.}
\label{fig:bernoulli_d}
\end{figure}

Figure \ref{fig:bernoulli_k} shows the performance of our method using various values of $K$ and $\eta$ applied to the $512$-bit \onemax{} and \leadingones{} problems.
Figure \ref{fig:bernoulli_d} shows the scaling-up of the performance with the problem dimension $d$ under $\eta = 1/d$. We observe that the effect of sample reuse is rather uniform with respect to dimension. 

In Figure~\ref{fig:bernoulli_k}, we see monotone improvement of the performance as we increase the number of iterations $K$ from which we reuse samples, except in the case $\eta = 16/d$ on \onemax{}. On \leadingones{}, some runs failed when $K = 0$ for $\eta \geq 4/d$. On the other hand, the proposed strategy ($K \geq 1$) succeeded in in all cases. Generally, the improvement from $K = 0$ to $K = 1$ is the greatest and is slight for further increases in $K$. Importantly, a large $K$ value did not \nnew{worsen} the performance, though the improvement was small. It is because an old distribution tends to be away from the current distribution, and the likelihood ratio for old samples tends to be very small, resulting in almost neglecting them.

For the case of $\eta = 16/d$ on \onemax{}, the performance with $K = 10$ is about $20$ percent worse than with the best setting ($K = 1$), but it is still better than $K = 0$. We conclude that the proposed strategy is robust with respect to the choice of $K$. A larger $K$ may be preferred in terms of the number of function evaluations, but we cost computational time and memory space proportional to $K$, to process and keep past samples. 




\section{Experimental Evaluation of the Sample Reuse CMA-ES}
\label{sec:expcma}
In this section, we evaluate the proposed sample reuse IGO with the Gaussian distribution on \new{the} standard continuous benchmark functions. We report the impact \nnew{on performance} of the number of sample reuse iterations $K$ and the population size $\lambda$ along with the scalability for dimension $d$ in Section~\ref{sec:exp_gaussian_reuse_igo}. In addition, we compare the proposed method with the pure rank-$\mu$ update CMA-ES with and without the importance mixing. Finally, in Section~\ref{sec:exp_gaussian_reuse_igo_hybrid} we demonstrate how to combine the proposed sample reuse with a component that is not derived in the IGO framework, namely the rank-one update of the covariance matrix.


\begin{table}[tbp]
\caption{The benchmark functions used in the experiment of the sample reuse CMA-ES, where $d$ indicates the problem dimension.}
\label{tbl:test_func_continuous}
\centering
\begin{tabular}{l|l}\hline
Name & Function definition \\ \hline \hline
Sphere     & $f(x) = \sum_{i=1}^{d} x_i^2$ \\
Ellipsoid	& $f(x) = \sum_{i=1}^{d} \left( 1000^{\frac{i-1}{d-1}} x_i \right)^2$ \\
Cigar	& $f(x) = x_1^2 + \sum_{i=2}^{d} \left( 1000 x_i \right)^2$ \\
Rosenbrock & $f(x) = \sum_{i=1}^{d-1} \left( 100 (x_{i+1} - x_i^2)^2 + (x_i - 1)^2 \right)$ \\
Ackley	& $f(x) = 20 - 20 \exp \left( -0.2 \sqrt{\frac{1}{d} \sum_{i=1}^d x_i^2} \right)$ \\
    & \qquad \qquad \qquad \qquad \qquad \qquad $+ e - \exp \left( \frac{1}{d} \sum_{i=1}^{d} \cos(2 \pi x_i) \right)$ \\
Bohachevsky	& $f(x) = \sum_{i=1}^{d-1} \left( x_i^2 + 2 x_{i+1}^2 - 0.3 \cos (3 \pi x_j) - 0.4 \cos (4 \pi x_{i+1}) + 0.7 \right)$ \\
Schaffer & $f(x) = \sum_{i=1}^{d-1} (x_i^2 + x_{i+1}^2)^{0.25} (\sin^2(50(x_i^2 + x_{i+1}^2)^{0.1}) + 1.0)$ \\
Rastrigin  & $f(x) = 10n + \sum_{i=1}^{d}\{ x_i^2 - 10 \cos(2\pi x_i) \}$  \\
\hline
\end{tabular}
\end{table}

\subsection{Benchmark Functions and General Setting}
Our benchmark set consists of the Sphere, Rosenbrock, Ellipsoid, Cigar, Ackley, Bohachevsky, Schaffer, and Rastrigin functions. \nnew{The definition of each benchmark function is shown in Table \ref{tbl:test_func_continuous}.}
The global optimum is located at $(1, \dots, 1)^\T$ for the Rosenbrock function and at $(0, \dots, 0)^\T$ for the other functions; the optimal value is $0$ \new{for all the functions}.

The initial mean vector $m^0$ of the Gaussian distribution is drawn uniform randomly from $[a, b]^d$ for each run, and the covariance matrix is initialized by $C^0 = \sigma^2 \eye$, where $\sigma = (b - a)/2$. The range is set to $[1, 5]^d$ for the Sphere, Ellipsoid, Cigar, and Rastrigin functions, $[-2, 2]^d$ for the Rosenbrock function, $[1, 30]^d$ for the Ackley function, $[1, 15]^d$ for the Bohachevsky function, and $[10, 100]^d$ for the Schaffer function.

Each run is terminated and regarded as \nnew{a} success \del{when}{}\new{if} the best objective value reaches smaller than $10^{-10}$.
\del{On the other hand, the}{}\new{Each} run is terminated and regarded as \nnew{a} failure \del{after}{}\new{if} $d \times 10^6$ function evaluations \new{are spent without finding the target function value}, or \del{when}{}\new{if} the minimal eigenvalue of the covariance matrix reaches $10^{-60}$ for the Schaffer function or $10^{-30}$ for the other functions. For each setting we conduct $50$ independent runs. We evaluate the performance of each method by the average number of function evaluations over successful runs divided by the success probability.

\nnew{Different values of} $\lambda$ and $K$ are tested in the experiments. For the Sphere, Rosenbrock, Ellipsoid, and Cigar functions, $\lambda \in \{4 + \lfloor 3 \ln d \rfloor, 0.2d, 0.4d, 0.8d, d, 2d, 4d, 8d, 16d\}$, \new{where the population size $4 + \lfloor 3 \ln d \rfloor$ is the default setting in the CMA-ES}. For the other multimodal functions, $\lambda = 2\big\lfloor 2^{i/2-1} \lambda_{\mathrm{base}} \big\rfloor$ for $i= 0, 1, \dots, 7$ with ${\lambda}_{\mathrm{base}} = 2\ln {d}, d, 2d, 10d$ for the Ackley, Bohachevsky, Schaffer, and Rastrigin functions, respectively. In preliminary experiments, we have observed failure with high probability when $\lambda \leq \ \lambda_{\mathrm{base}}$. The number of iterations for sample reuse is $K \in \{0, 1, 3, 5, 7, 9\}$. As the utility function used in the proposed method is different \del{with}{}\new{from} the one in the standard CMA-ES, $K = 0$ leads to parameter updates slightly different from the standard mean vector update and the rank-$\mu$ update.\footnote{\new{The utility function used in \citep{Shirakawa2015}\nnew{, which is derived based on \eqref{eq:igo_utility},} is also slightly different \nnew{from the one }in this paper\del{, which is derived based on \eqref{eq:igo_utility}}{}. The total value of the resulting weights ($\sum \hat{w}_i / \lambda$) is almost always less than $1$ in \citep{Shirakawa2015}, while the sum of the weights derived from \eqref{eq:weight_estimated} and \eqref{eq:cmaw} almost always \nnew{equals to} $1$.}} We chose the learning rate based on the CMA-ES default parameters as follows:
\begin{equation}
\cm = 1, \quad \cmu = \frac{2 (\mueff - 2 + 1/\mueff)}{(d + 2)^2 + 2 \mueff/2},
\label{eq:lr_reuse}
\end{equation}
where $\mueff$ is calculated using \eqref{eq:cma-weight}.

\subsection{Sample Reuse in the Pure Rank-$\mu$ Update CMA-ES}
\label{sec:exp_gaussian_reuse_igo}

We compare the following two variants for the proposed method:
\begin{description}
\setlength{\itemsep}{0em}
\item[(A) Reuse-$m$, $C$] The parameters are updated using  \eqref{eq:ngm} and \eqref{eq:ngc}.
\item[(B) Reuse-$C$] The covariance matrix is updated using \eqref{eq:ngc}, while the mean vector is updated by \eqref{eq:igo-m-update}, \new{i.e., }only current samples are used, and the rankings are computed among the current samples as in the standard CMA-ES.
\end{description}

\begin{figure}[tbp]
  \newcommand\figsize{0.99\linewidth}
  \centering
  \includegraphics[width=\figsize]{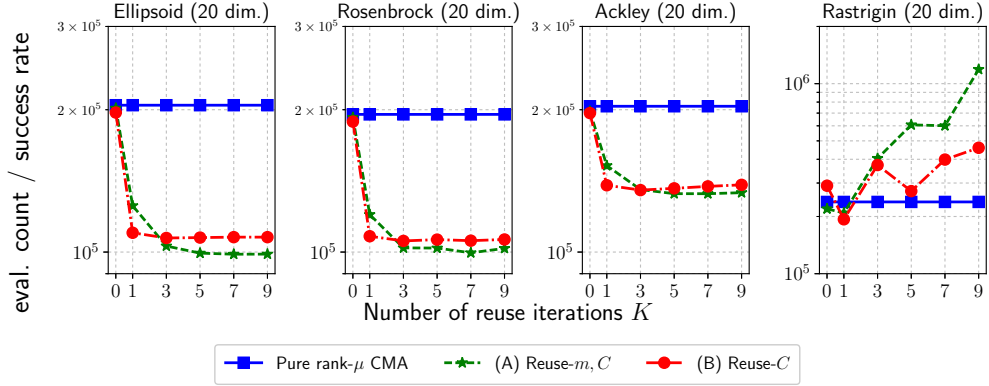}
  \caption{The search performance of the pure rank-$\mu$ update CMA-ES, the algorithms (A) and (B) for various $K$ on $20$-dimensional problems. The default population size $4 + \lfloor 3 \ln d \rfloor$ is used for \nnew{Ellipsoid and} Rosenbrock functions, and $\lambda = 2\big\lfloor 2 \lambda_{\mathrm{base}} \big\rfloor$ for Ackley and Rastrigin functions.}
  \label{fig:perform_reuseAB_k_20}
\end{figure}
\begin{figure}[tbp]
  \newcommand\figsize{0.99\linewidth}
  \centering
  \includegraphics[width=\figsize]{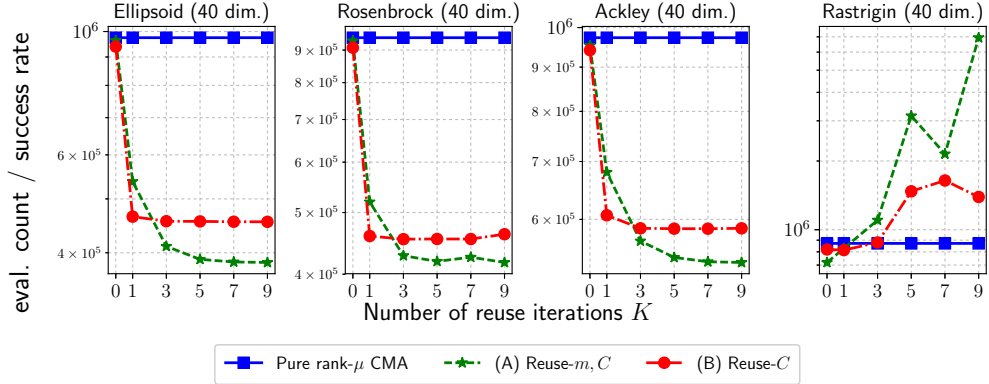}
  \caption{The search performance of the pure rank-$\mu$ update CMA-ES, the algorithms (A) and (B) for various $K$ on $40$-dimensional problems. The default population size $4 + \lfloor 3 \ln d \rfloor$ is used for \nnew{Ellipsoid and} Rosenbrock functions, and $\lambda = 2\big\lfloor 2 \lambda_{\mathrm{base}} \big\rfloor$ for Ackley and Rastrigin functions.}
  \label{fig:perform_reuseAB_k_40}
\end{figure}

Figure\new{s}~\ref{fig:perform_reuseAB_k_20} and \ref{fig:perform_reuseAB_k_40} show the search performances for \new{different} $K$ \new{values} on $20$- and $40$-dimensional functions, respectively. The results using the default population size $\lambda = 4 + \lfloor 3 \ln d \rfloor$ is shown for the \nnew{Ellipsoid and Rosenbrock} functions, and $\lambda = 2\big\lfloor 2 \lambda_{\mathrm{base}} \big\rfloor$ for the Ackley and Rastrigin functions.
The results for the \nnew{Sphere and Cigar functions were} very similar to those for the \nnew{Ellipsoid} function, and the results for the Bohachevsky and Schaffer functions were similar to those for the Ackley function. \del{Moreover, w}{}\new{W}e observed \del{a}{}similar trends with respect to $K$ on the $40$-dimensional functions.

\begin{figure}[tb]
\newcommand{\myfigsize}{0.99\linewidth}
\newcommand{\mysubfigsize}{0.32\linewidth}
\centering
\begin{subfigure}[]{\mysubfigsize}
  \centering  
  \includegraphics[width=\myfigsize]{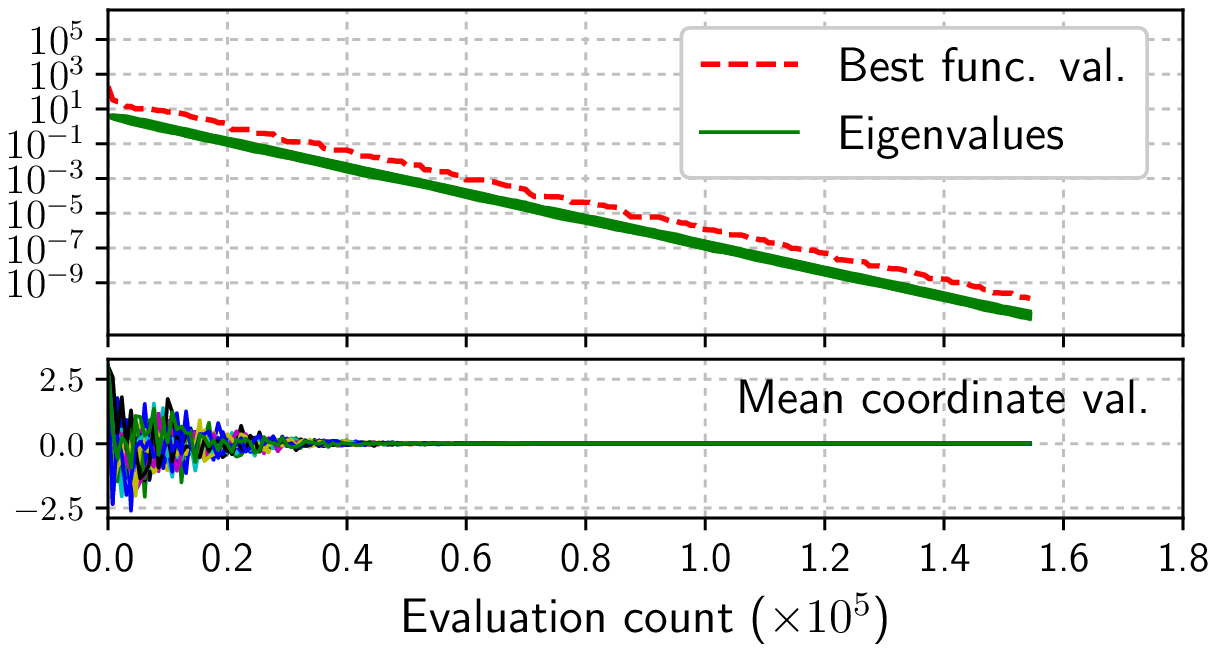}
  \subcaption{Sphere ($K=0$)}
\end{subfigure}
\begin{subfigure}[]{\mysubfigsize}
  \centering  
  \includegraphics[width=\myfigsize]{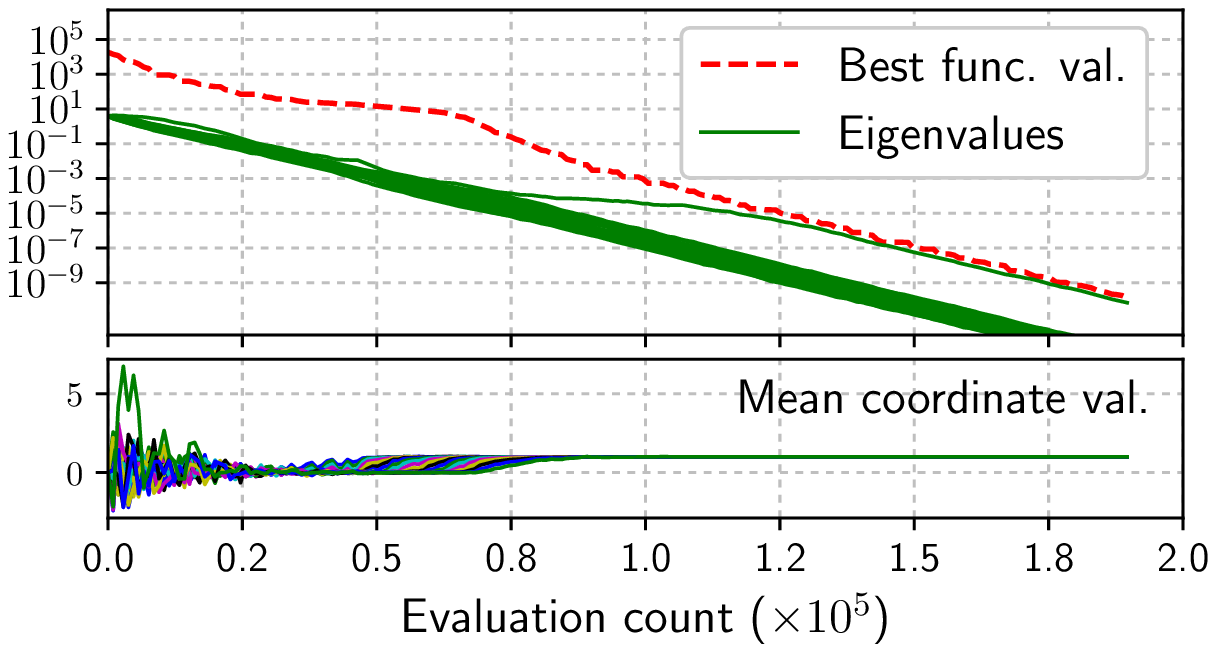}
  \subcaption{Rosenbrock ($K=0$)}
\end{subfigure}
\begin{subfigure}[]{\mysubfigsize}
  \centering  
  \includegraphics[width=\myfigsize]{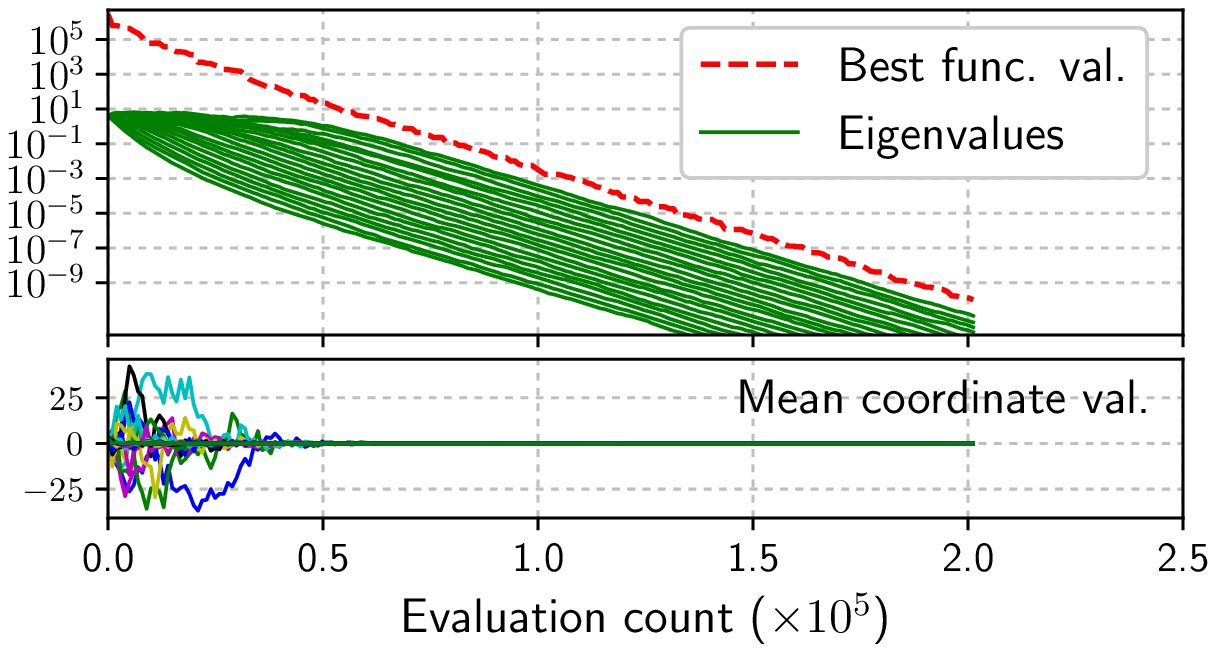}
  \subcaption{Ellipsoid ($K=0$)}
\end{subfigure}
\begin{subfigure}[]{\mysubfigsize}
  \centering  
  \includegraphics[width=\myfigsize]{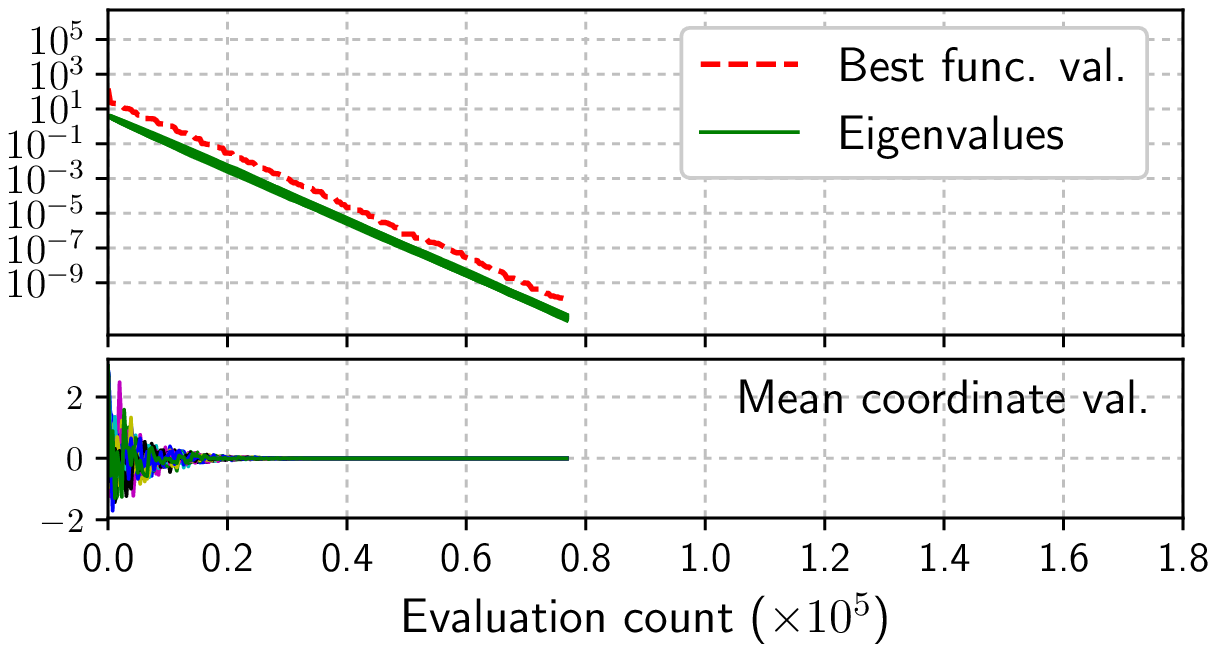}
  \subcaption{Sphere ($K=5$)}
\end{subfigure}
\begin{subfigure}[]{\mysubfigsize}
  \centering  
  \includegraphics[width=\myfigsize]{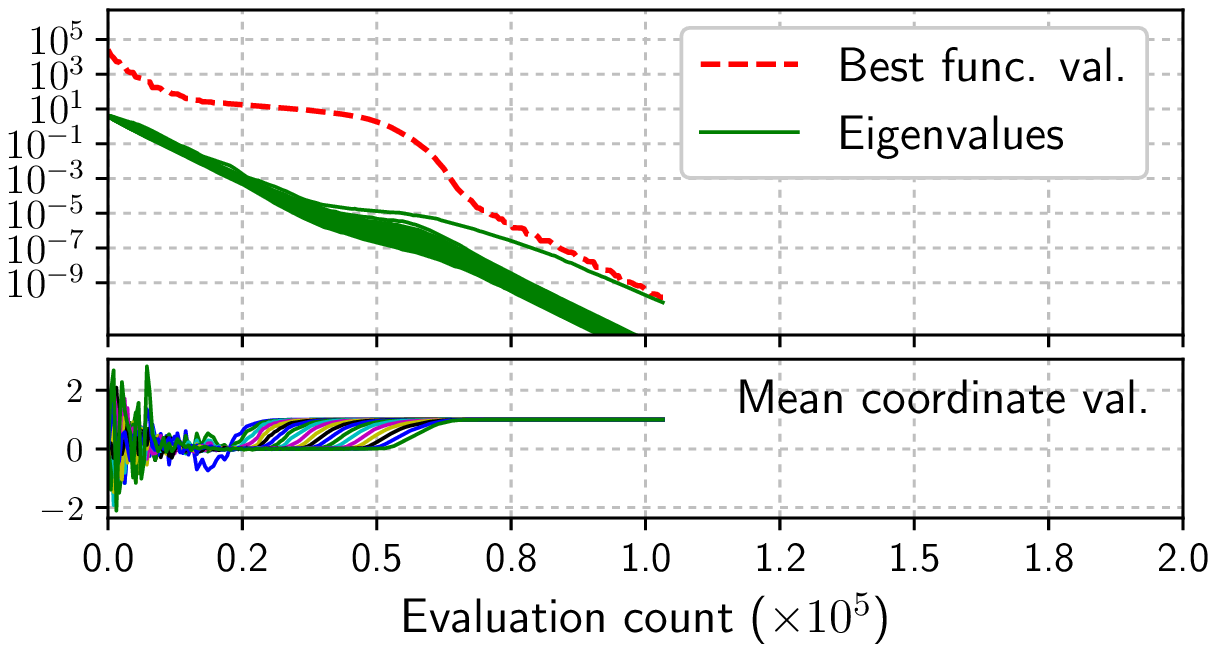}
  \subcaption{Rosenbrock ($K=5$)}
\end{subfigure}
\begin{subfigure}[]{\mysubfigsize}
  \centering  
  \includegraphics[width=\myfigsize]{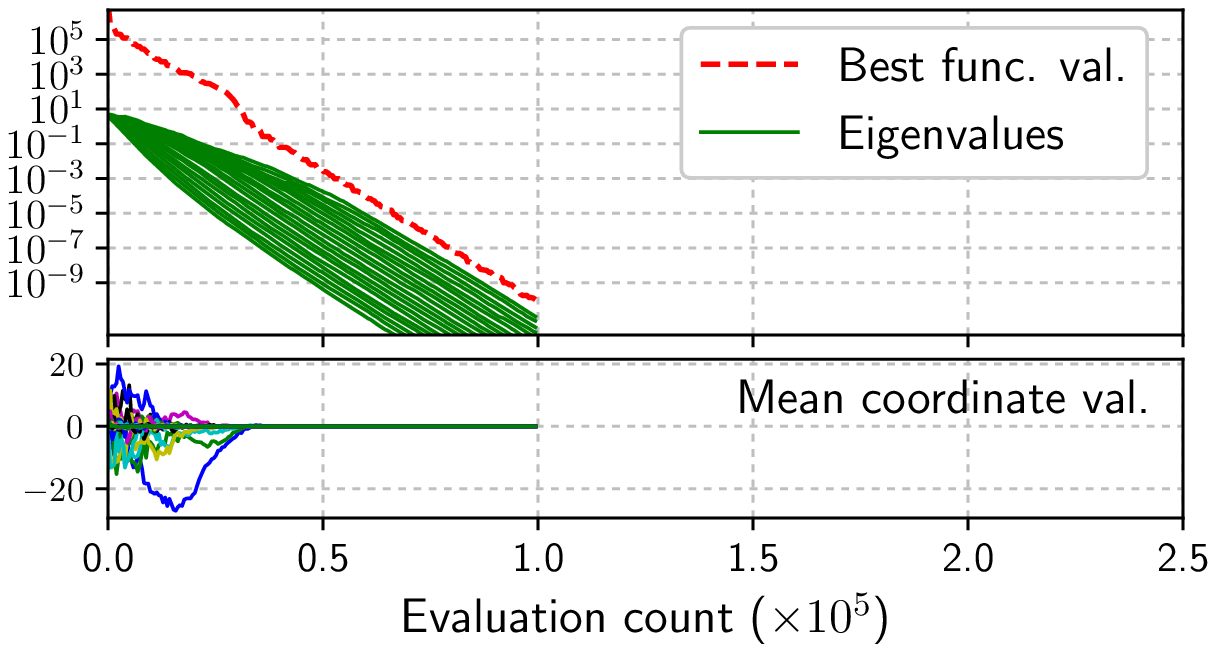}
  \subcaption{Ellipsoid ($K=5$)}
\end{subfigure}
\caption{The best objective function value, eigenvalues of $C$, and each coordinate of $m$ with $K=0$ (top) and $K=5$ (bottom) are shown for the algorithm (A) Reuse-$m, C$ with the default population size on $20$-dimensional Sphere (1st column), Rosenbrock (2nd column), and Ellipsoid (3rd column) functions.
}
\label{fig:eigenvalue_a}
\end{figure}

Figure \ref{fig:eigenvalue_a} displays the typical behaviors of the algorithm (A) with the default population size on $20$-dimensional Sphere, Rosenbrock, and Ellipsoid functions. The results of both $K = 0$ and $K = 5$ are monitored.

\begin{figure}[tb]
\centering
\begin{subfigure}[]{0.49\linewidth}
  \centering  
  \includegraphics[width=0.99\linewidth]{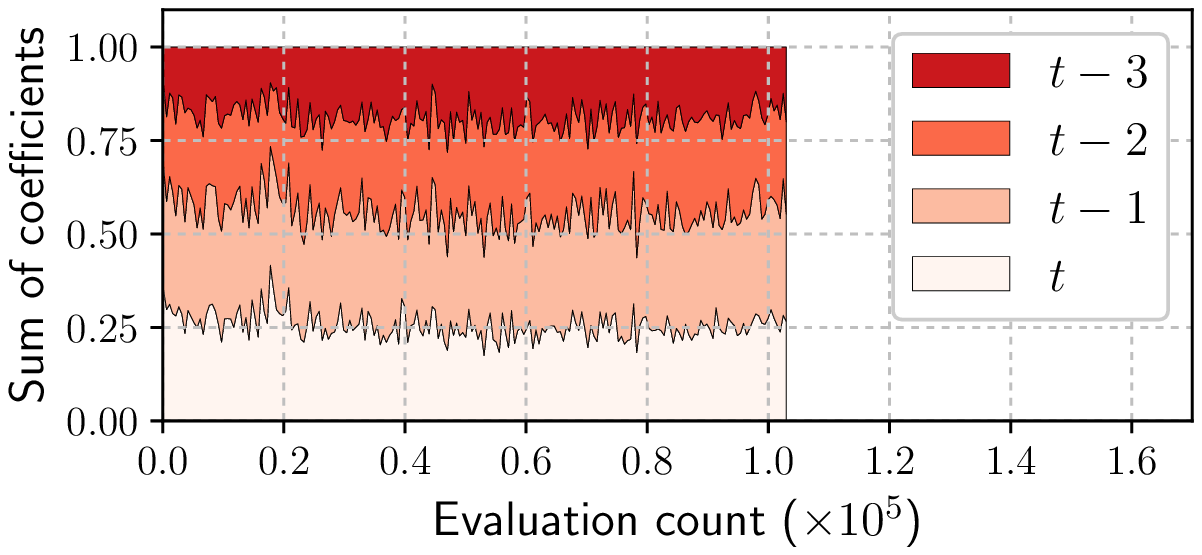}
  \subcaption{Algorithm (A) Reuse-$m, C$}
\end{subfigure}
\begin{subfigure}[]{0.49\linewidth}
  \centering  
  \includegraphics[width=0.99\linewidth]{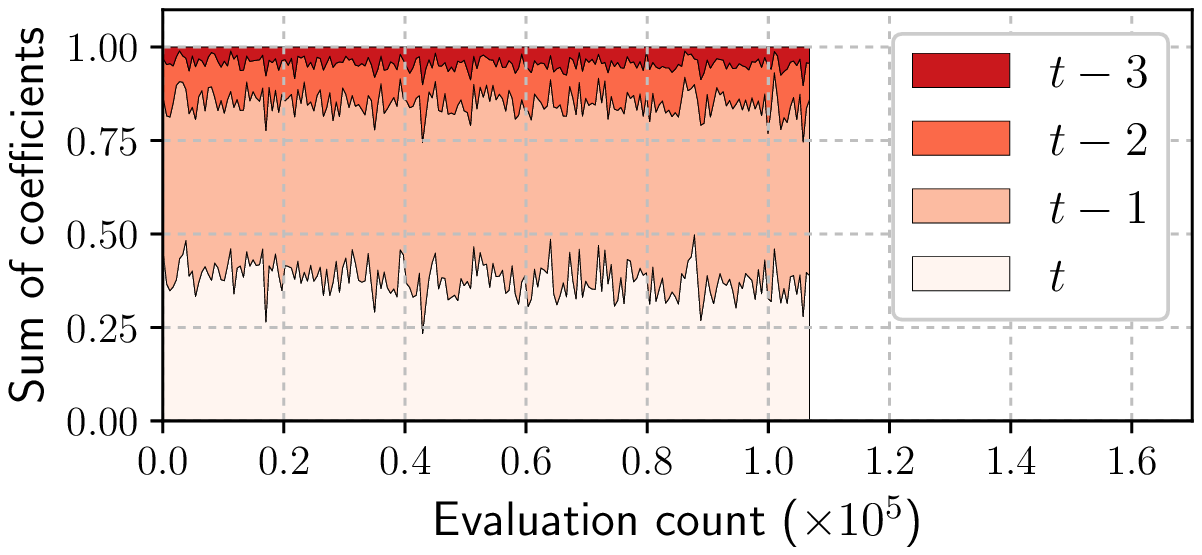}
  \subcaption{Algorithm (B) Reuse-$C$}
\end{subfigure}
\caption{Transition of the sum of the coefficients (the products of the weight and the likelihood ratio) for the current and past samples, $s_t^k = 1/\lambda \sum_{i} \hat{r}(x^k_i), k=t, \dots, t-3$, on $20$-dimensional Ellipsoid function for the algorithms (A) and (B) with $K=3$ and the default population size. The moving average with window size $11$ has been applied for smoothing.}
\label{fig:weight_sum_ellipsoid}
\end{figure}

Figure \ref{fig:weight_sum_ellipsoid} illustrates the transition of the sum of the products of the weight and the likelihood ratio $s_t^k = (1/\lambda) \sum_{i} \hat{r}(x^k_i)$ for the current ($k = t$) and past samples ($k = t-1, \dots, t-3$) in a typical run. The results of the algorithm\nnew{s} (A) and (B) with $K = 3$ and the default population size on $20$-dimensional Ellipsoid function are shown. Smaller values for \nnew{past} iterations indicate either that the samples from that iteration are away from the current distribution and the likelihood ratios are small, or that the samples from that iteration are low in the ranking and small or zero weight values have been assigned to them.

\begin{figure}[tbp]
\centering
\begin{subfigure}[]{0.44\linewidth}
  \centering  
  \includegraphics[height=4.2cm]{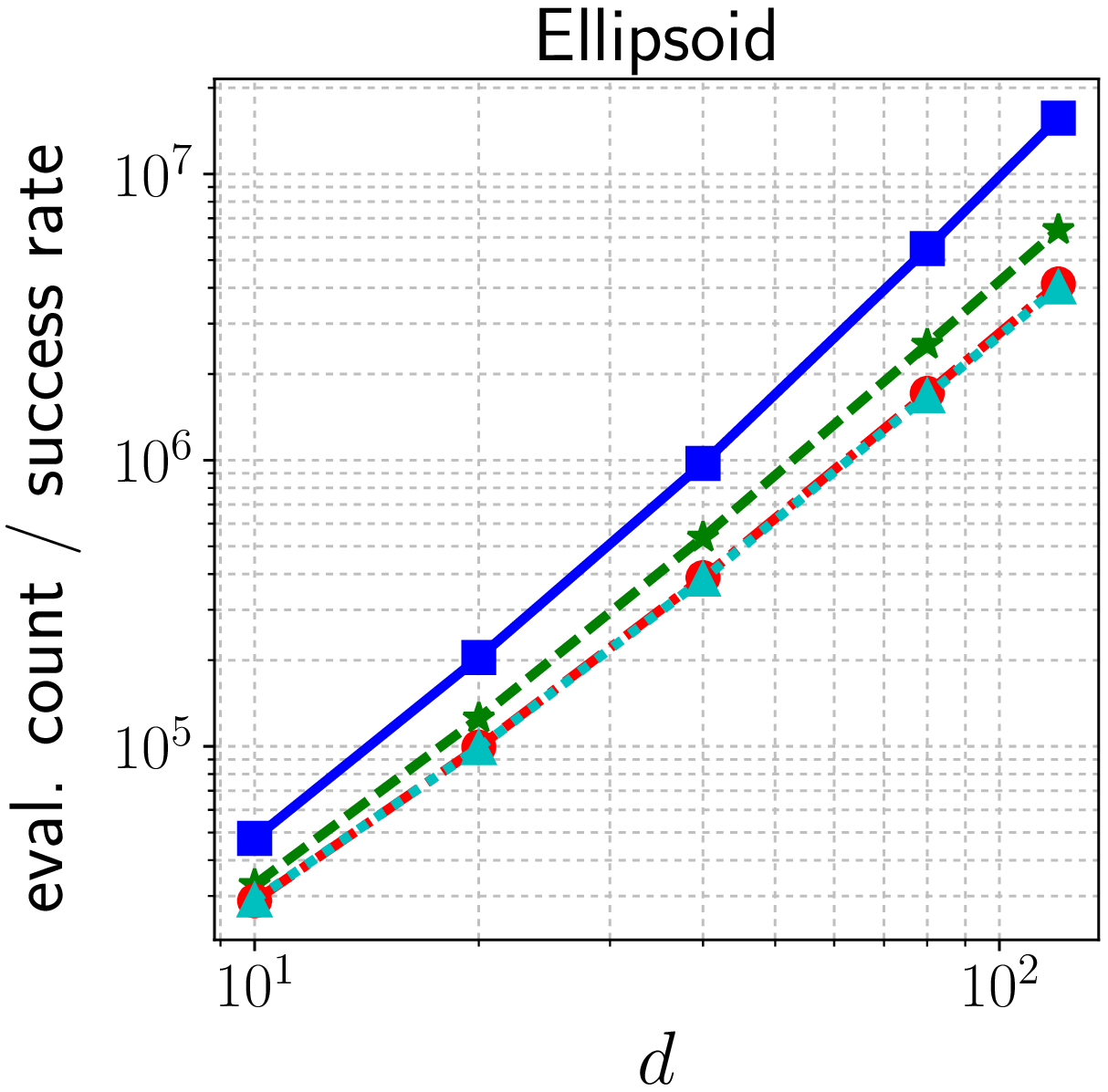}
\end{subfigure}
\begin{subfigure}[]{0.54\linewidth}
  \centering  
  \includegraphics[height=4.2cm]{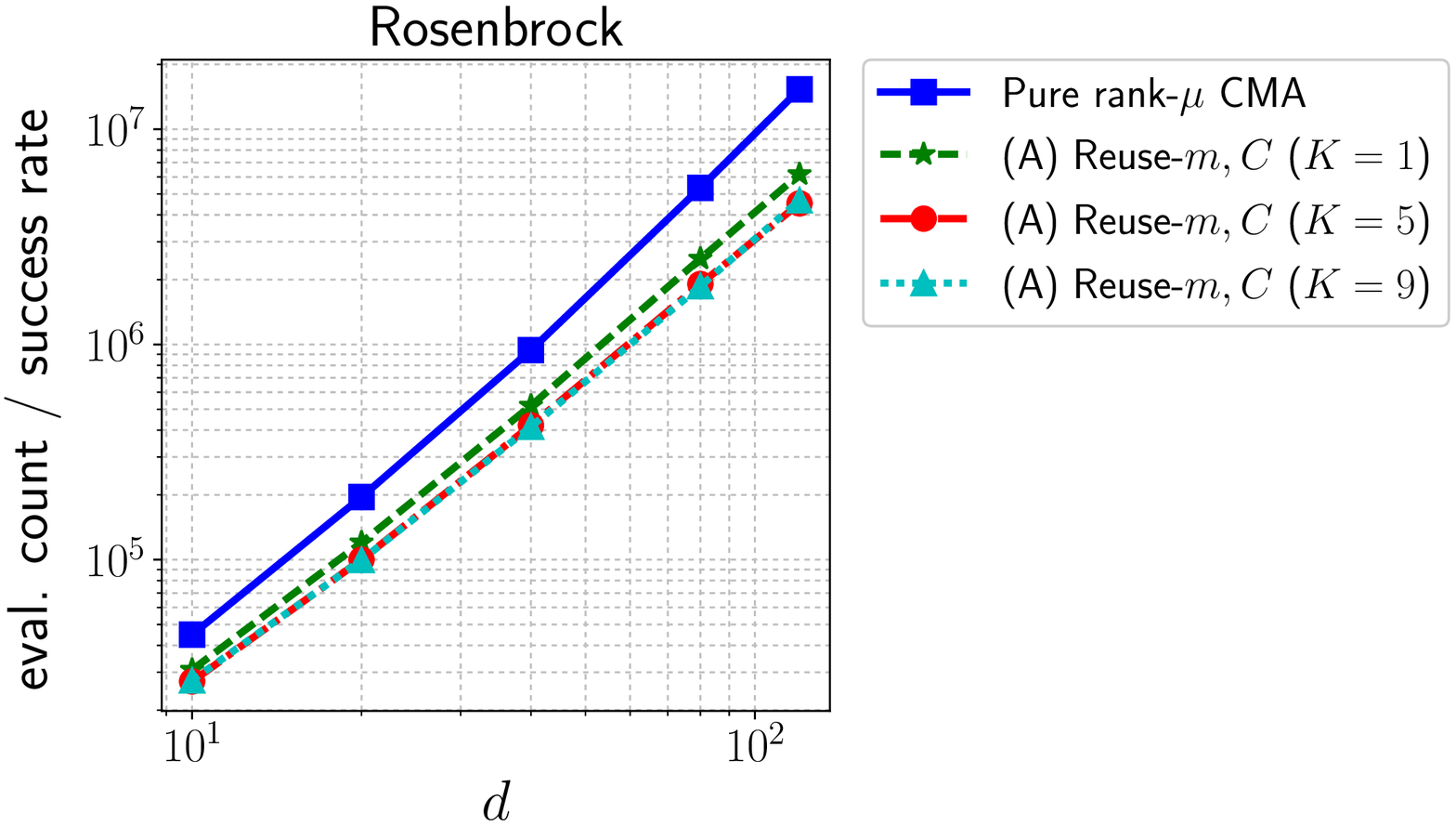}
\end{subfigure}
\caption{The search performance of (A) Reuse-$m, C$ for the problem dimension $d$. The results on the Ellipsoid (left) and Rosenbrock (right) functions are displayed against the different numbers of sample reuse iterations ($K \in \{1, 5, 9\}$).}
\label{fig:gaussian_d}
\end{figure}

\new{Figure \ref{fig:gaussian_d} shows the performances of \nnew{the algorithm} (A) \del{Reuse-$m, C$}{} and the pure rank-$\mu$ update CMA-ES for different problem dimensions. The results using the numbers of sample reuse $K \in \{ 1, 5, 9\}$ on Ellipsoid and Rosenbrock functions are reported.} \nnew{The effect of sample reuse is more pronounced for a larger dimension $d$ in this setting. }

\begin{figure}[tbp]
  \newcommand\figsize{0.99\linewidth}
  \centering
  \includegraphics[width=\figsize]{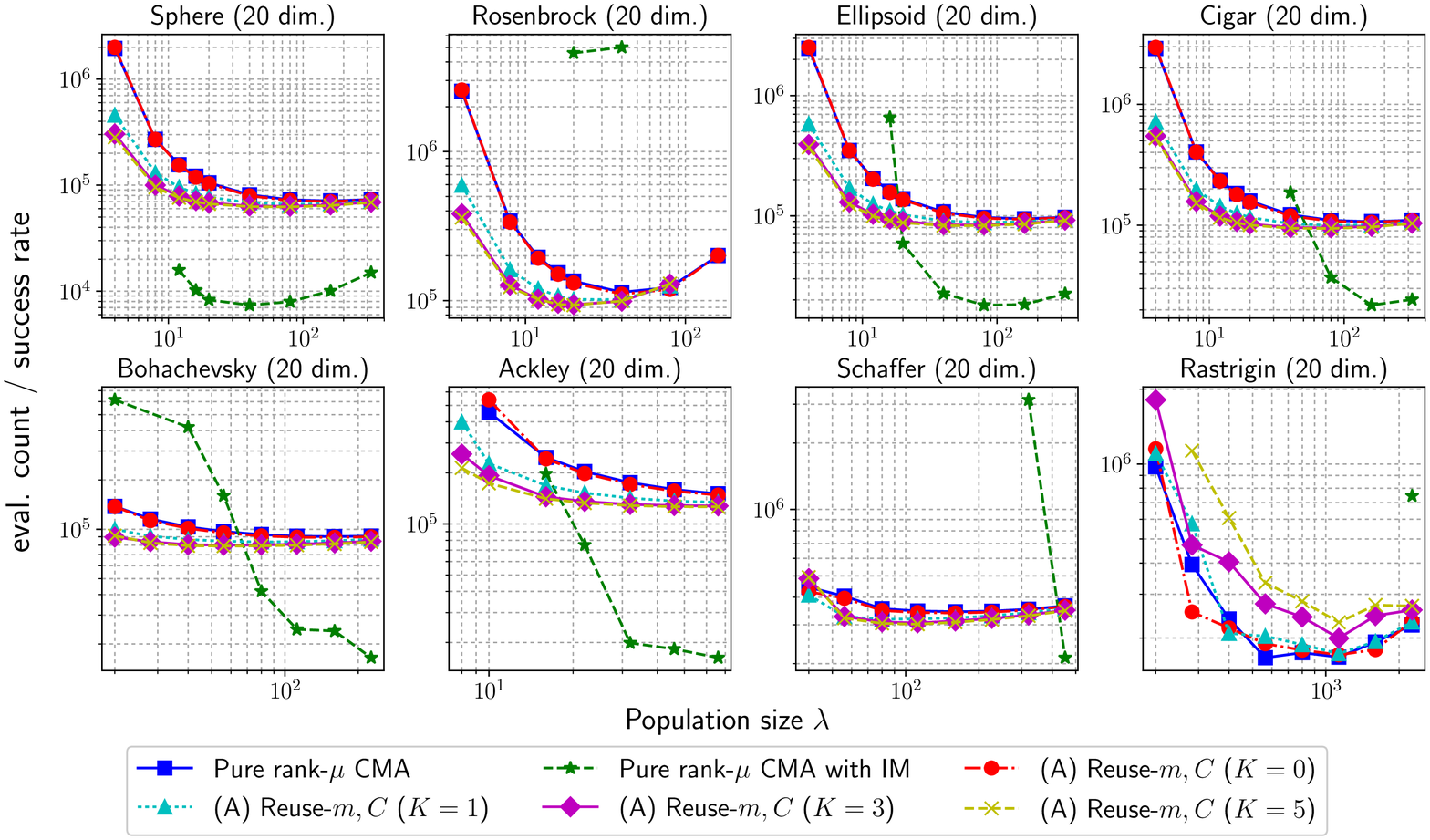}
  \caption{The search performance versus population size $\lambda$ of the pure rank-$\mu$ update CMA-ES with and without importance mixing (IM), and the rank-$\mu$ update CMA-ES with sample reuse (the algorithm~(A) for $K \in \{0, 1, 3, 5\}$) on $20$-dimensional problems. Missing data implies the failure.}
  \label{fig:perform_reuseA_20}
\end{figure}

\begin{figure}[tbp]
  \newcommand\figsize{0.99\linewidth}
  \centering
  \includegraphics[width=\figsize]{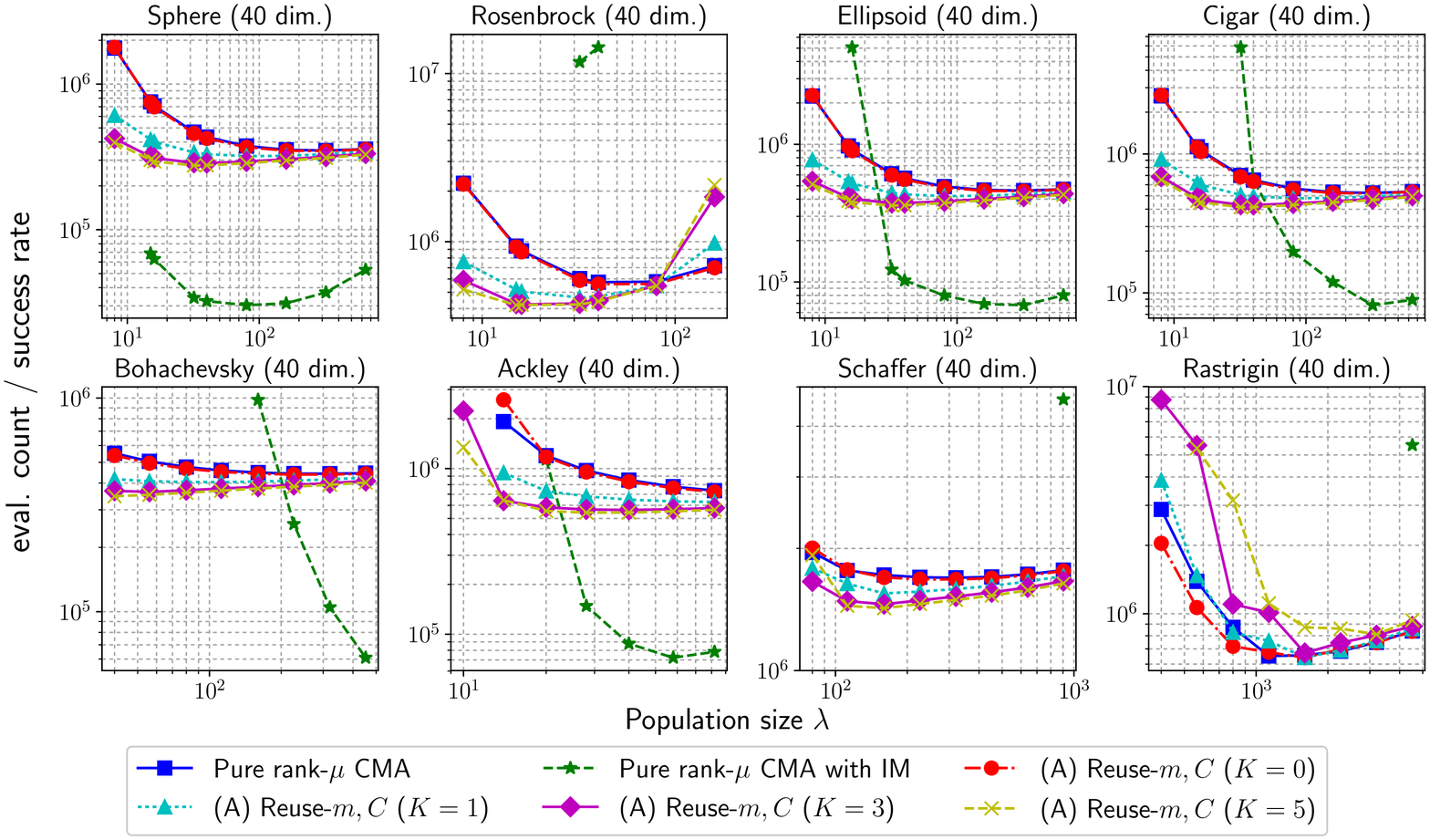}
  \caption{The search performance versus population size $\lambda$ of the pure rank-$\mu$ update CMA-ES with and without importance mixing (IM), and the rank-$\mu$ update CMA-ES with sample reuse (the algorithm~(A) for $K \in \{0, 1, 3, 5\}$) on $40$-dimensional problems. Missing data implies the failure.}
  \label{fig:perform_reuseA_40}
\end{figure}

Figures \ref{fig:perform_reuseA_20} and \ref{fig:perform_reuseA_40} show the search performance\nnew{s} for varying population size on $20$ and $40$-dimensional problems, respectively. The performances of the pure rank-$\mu$ update CMA-ES, the pure rank-$\mu$ update CMA-ES with importance mixing, and the pure rank-$\mu$ update CMA-ES with sample reuse \nnew{(algorithm (A) Reuse-$m, C$)} \nnew{are shown for different population sizes, with $K \in \{0, 1, 3, 5\}$}. \new{The result for $40$-dimensional problems had similar trends as for $20$-dimensional problems.}



\paragraph{Impact of $K$}
We observe from Figures~\ref{fig:perform_reuseAB_k_20} and \ref{fig:perform_reuseAB_k_40} that the performance of the sample reuse CMA-ES improves more or less monotonically as $K$ increases for all but the Rastrigin function. The sample reuse CMA-ES outperforms the one without sample reuse (i.e., the pure rank-$\mu$ update CMA-ES) already with $K = 1$, where the current and previous populations are used. We see the improvement as $K$ increases up to $7$ \new{in the algorithm (A)}. The performance is saturated for large $K$ because the likelihood ratios for old samples tend to be small, since past distributions are away from the current distribution and newer samples tend to have better function values, resulting that very small weights are assigned to very old samples. On the Rastrigin function, a large $K$ value tends to lead to low success rate. From Figure~\ref{fig:eigenvalue_a}, the speedup by the sample reuse ($K = 5$ over $K = 0$) is seen not only at the convergence stage but also at the adaptation stage. Comparing the algorithms (A) and (B) in Figures~\ref{fig:perform_reuseAB_k_20} and \ref{fig:perform_reuseAB_k_40}, we find that the algorithm (B) is better for $K = 1$ but the algorithm (A) reaches better performance for larger $K$. The reason is observed in Figure~\ref{fig:weight_sum_ellipsoid}, where the sums of the coefficients $\hat{r}(x^k_i)$, which appear in \eqref{eq:ngm} and \eqref{eq:ngc}, are more or less equal for $k = t, \dots, t - 3$ in the algorithm (A), whereas the sums for $k = t$ and $t - 1$ amount to higher values than the ones for $k = t - 2$ and $t - 3$ in the algorithm (B). This indicates that the past samples are being used to estimate the natural gradient in the algorithm (A) as well as the current samples, whereas mostly the current and previous samples are being used, the older samples \del{has}{}\nnew{have} less impact in the algorithm (B). In the algorithm (A), the mean vector is updated by using the proposed update \eqref{eq:ngm}, where the natural gradient is estimated with smaller variance than the one estimated in the original update \eqref{eq:igo-m-update}. Then, the mean vector less fluctuates, and the likelihood ratios are likely to be relatively large for the past samples. \new{However, we note that the bad influence of the sample reuse on the Rastrigin function decreases in the algorithm (B).}

\new{The importance sampling itself \del{do}\nnew{does} not add any bias in the natural gradient estimate. However, since the current distribution parameter is the result of its update using the past samples, the current samples that depend\del{s}{} on the current parameters and the past samples that are used in the previous parameter update are somewhat correlated. These correlation\nnew{s} may not be negligible if $K \geq 1$. This might be the reason \del{of}{}\nnew{for} the performance deterioration on the Rastrigin and Rosenbrock functions.}

\paragraph{Impact of $\lambda$}
When the population size is relatively large, we observe from Figures~\ref{fig:perform_reuseA_20} and \ref{fig:perform_reuseA_40} that the sample reuse is not helpful very much to reduce the number of function evaluations. For large $\lambda$, we expect that the natural gradient estimate in \eqref{eq:igo-m-update} and \eqref{eq:igo-c-update} is relatively accurate and the effect of the sample reuse is less emphasized. Moreover, since the learning rate $\cmu$ defined in \eqref{eq:lr_reuse} becomes larger as $\lambda$ increases, leading to a big change of the probability distribution, the likelihood ratios for the past samples will be small. Therefore, the sample reuse does not help when $\lambda$ is relatively large. We can observe the same effect of large $\lambda$ for the algorithms (B) Reuse-$C$.

\del{On Rosenbrock function, the sample reuse turns out to lead to failure in the case of $\lambda > 10^{2}$. 
As we see in Figure~\ref{fig:eigenvalue_a}, the mean vector needs to move from the origin towards the global optimum. In the case of $\lambda = 160$ and $K = 5$, we observe that the eigenvalues of the covariance matrix tends to be smaller compared to the case of $K = 0$ during the movement of the mean vectors. Smaller eigenvalues of the covariance matrix leads to producing shorter steps, $x - m$, resulting in slow movement of the mean vectors.\footnote{Empirically, we observe that a larger population size itself leads to smaller eigenvalues. Since more samples are used in the case of $K = 5$, compared to the case of $K = 0$, the reason of adapting a smaller covariance matrix may be the same reason. } }


\paragraph{Comparison with importance mixing}
When introducing the importance mixing in the pure rank-$\mu$ update CMA-ES, \new{the runs with small population size \nnew{tends to }fail or deteriorate}\del{the performance on the unimodal functions are improved}{}. \new{The reason is that the most of the samples become recycled samples when the population size is small, thereby the distribution did not converge.} \new{In addition, the Rosenbrock, Schaffer, and Rastrigin functions cannot be solved with most population sizes when introducing the importance mixing.} \new{While the performances on the Sphere, Ellipsoid, Cigar, Bohachevsky, and Ackley functions improve}\del{The improvement is emphasized}{} when $\lambda$ is relatively large, as is seen in Figures~\ref{fig:perform_reuseA_20} and \ref{fig:perform_reuseA_40}. \del{The performance is better than the algorithm (A) for $\lambda \geq 320$. On the other hand, it is likely to be trapped by local minima on the multimodal functions.}{}The motivation of the importance mixing is to reduce the number of function evaluations by reusing the samples and not to estimate the natural gradient more precisely. In fact, the estimate of the natural gradient will be less accurate since the number of samples drawn from the current distribution is less than $\lambda$. 

\del{Combining the rank-one update and the importance mixing, it ceases working properly when the refresh rate $\alpha = 0$. By tuning $\alpha$, the unimodal functions can be solved. However, the $\alpha$ needs to be tuned dependently of $\lambda$. Moreover, it does not work well on the multimodal functions.}{}

\shin{The implementation of importance mixing in \cite{Shirakawa2015} was incorrect. It used $y = x_{t-1} - m_{t-1}$ to compute the natural gradient for the accepted past samples instead of $y = x_{t-1} - m_{t}$.}

\subsection{Sample Reuse in the Hybrid Update}
\label{sec:exp_gaussian_reuse_igo_hybrid}

Here we incorporate the rank-one update into \eqref{eq:ngc}. We simply add the second term on the RHS of \eqref{eq:hybrid} to \eqref{eq:ngc}. To update the evolution path, we solely use the current samples and use \nnew{\eqref{eq:cma-weight}} as the weight for each $x_i^{(t)}$ rather than \nnew{\eqref{eq:weight_estimated}}. It means the evolution path is updated as it is done in \eqref{eq:evolution-path}. Since the evolution path itself accumulates the past information, the rank-one update is considered utilizing the past samples. Therefore, the proposed algorithm exploits the past information in different ways to update the covariance matrix, and we expect the synergy.
\new{Our objective here is to demonstrate the flexibility of the proposed sample reuse technique in the sense that we can combine it with a component that is not derived from the IGO framework.}

We consider the following two variants of the combination with the rank-one update:
\begin{description}
\item[(C) Reuse-$m$, $C$ $+$ rank-one] The rank-one update is incorporated into \eqref{eq:ngc}, and the mean vector is updated using the past samples as in \eqref{eq:ngm}.
\item[(D) Reuse-$C$ $+$ rank-one] The rank-one update is incorporated into \eqref{eq:ngc}, while the mean vector is updated by \eqref{eq:igo-m-update}. Only current samples are used, and the rankings are computed among the current samples as in the standard CMA-ES.
\end{description}

For these two variants, we observe the effect of $K$ and the population size $\lambda$. Moreover, we compare them with the standard hybrid of the rank-one and rank-$\mu$ update.
Following \citep{Hansen2014}, we set the learning rates as follows:
\begin{equation}
\begin{split}
\cm &= 1, \quad
\cc = \frac{4 + \mueff / d}{\lambda + 4 + 2 \mueff /\lambda}, \quad\cone = \frac{2}{(d + 1.3)^2 + \mueff}, \\
\cmu &= \min \left( 1.0-c_1, \frac{2 (\mueff - 2 + 1/\mueff)}{(d + 2)^2 + 2 \mueff/2} \right).
\end{split}\label{eq:rl}
\end{equation}
The evolution path $\pc$ is initialized to the zero vector.

\begin{figure}[tbp]
  \newcommand\figsize{0.99\linewidth}
  \centering
  \includegraphics[width=\figsize]{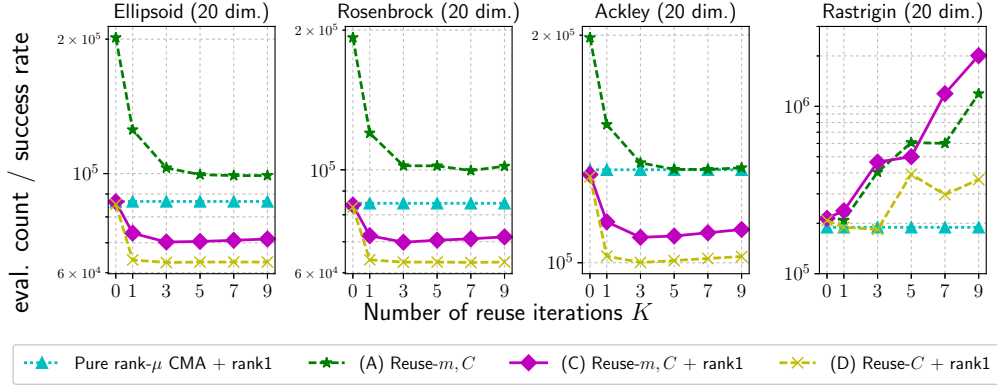}
  \caption{The search performance of the pure rank-$\mu$ CMA-ES \nnew{with the} rank-one update, the algorithms (A), (C), and (D), for $K$ on $20$-dimensional problems. The default population size is used for Ellipsoid and Rosenbrock functions and $\lambda = 2\big\lfloor 2 \lambda_{\mathrm{base}} \big\rfloor$ for Ackley and Rastrigin functions.}
  \label{fig:perform_k_20_CD}
\end{figure}

\begin{figure}[tbp]
  \newcommand\figsize{0.99\linewidth}
  \centering
  \includegraphics[width=\figsize]{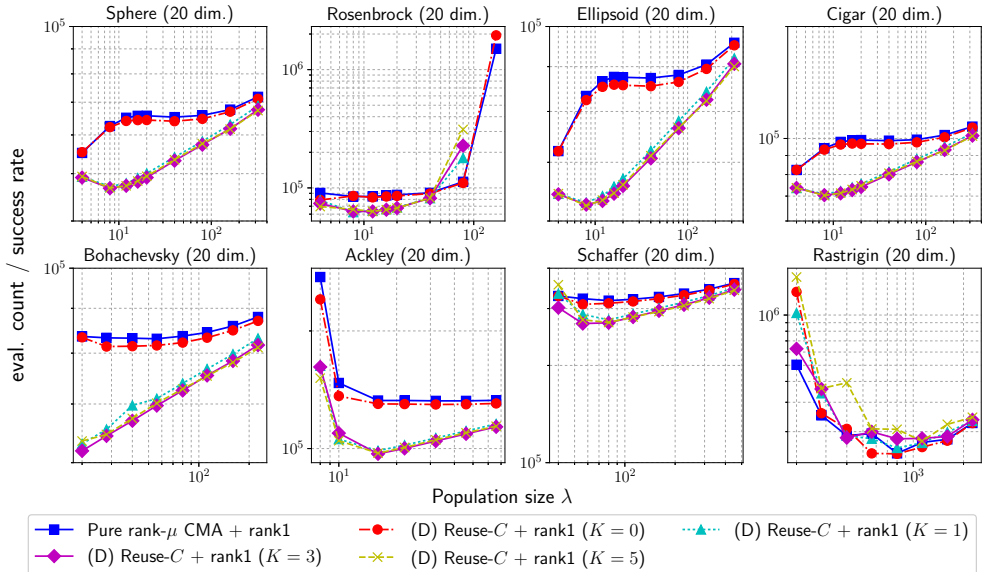}
  \caption{The search performance versus population size $\lambda$ of the pure rank-$\mu$ update CMA-ES with the rank-one update and the CMA-ES with sample reuse (the algorithm~(D) for $K \in \{0, 1, 3, 5\}$) on $20$-dimensional problems.}
  \label{fig:perform_reuseCrank1_20}
\end{figure}

Figure \ref{fig:perform_k_20_CD} shows the average number of function evaluations over the successful runs divided by the success probability on $20$-dimensional functions for \del{three}{}\nnew{four} algorithms: the hybrid update CMA-ES (described in Section~\ref{sec:cma-es}), the algorithms (A), (C), and (D). The default population size $\lambda = 4 + \lfloor 3 \ln d \rfloor$ is used for the \nnew{Ellipsoid and} Rosenbrock\del{, and Cigar}{} functions, and $\lambda = 2\big\lfloor 2 \lambda_{\mathrm{base}} \big\rfloor$ for the Ackley and Rastrigin functions. The result for the Ellipsoid function was very similar to that for the Cigar function, and the results for the Bohachevsky and Schaffer functions were similar to that for the Ackley function. Moreover, we observed similar trends with respect to $K$ on $40$-dimensional functions. 

Introducing the rank-one update improves the performance as seen in Figure~\ref{fig:perform_k_20_CD}. While the sample reuse for $m$ update results in better performance when the covariance matrix is solely updated by the rank-$\mu$ update, the standard $m$ update leads to faster convergence when the rank-one update is incorporated. As discussed in Section~\ref{sec:exp_gaussian_reuse_igo}, the proposed $m$ update \eqref{eq:ngm} tends to have less fluctuation in a subspace that \nnew{is} less sensitive \del{in}{}\nnew{to} function value, such as the first axis for the Cigar function. Then the evolution path will not be long in this subspace, while it should be in the example of the Cigar function so that the covariance matrix learns the principle axis. The standard $m$ update does not disturb the rank-one update, and the speedup is achieved by the sample reuse in the rank-$\mu$ update. 

Comparing the performance on the Rastrigin function shown in Figures~\ref{fig:perform_k_20_CD}, we notice that the algorithm (D) is less suffered from a large value of $K$ than the algorithm (C). This may be because the past samples have less effect in the algorithms (B) and (D), for the reason partially observed in Figure~\ref{fig:weight_sum_ellipsoid}.

Figure \ref{fig:perform_reuseCrank1_20} shows the \del{average number of function evaluations over the successful runs divided by the success probability}{}\nnew{search performance} on $20$-dimensional problems with varying population size. The results of the hybrid update CMA-ES and the algorithm (D) Reuse-$C$ + rank-one with $K \in \{ 0, 1, 3, 5 \}$ are plotted.
\del{Figure \ref{fig:perform_reuseCrank1_20} shows}{}\nnew{We observe} that the smallest $\lambda$ leads to the least performance difference between different $K$ values in the algorithm (D) for the unimodal functions, while the difference between different $K$ values becomes pronounced for smaller $\lambda$ in the algorithm (A), as shown in Figure~\ref{fig:perform_reuseA_20}. The learning rate $\cone$ for the rank-one update has a relatively large value for a small population size compared with the learning rate $\cmu$ for the rank-$\mu$ update, which means the impact of the rank-one update is dominative for a small population size.

\section{Conclusion}
\label{sec:conclusion}
We proposed a sample reuse technique based on the importance sampling in the recently proposed generic framework for probability model-based search algorithms, namely information geometric optimization framework. By reusing previously generated and evaluated samples to improve the accuracy of the estimate of the natural gradient in the IGO, we reduce the total number of function evaluations, which is often the bottleneck of the optimization process. The proposed sample reuse technique tries to improve the accuracy of the natural gradient estimate without affecting any working principles of the IGO framework. This means the technique can be applied to any instantiation of the IGO algorithms, including the well-known PBIL and the CMA-ES, and future algorithms derived from the IGO framework. Moreover, it can be basically combined with other techniques that seek to improve the IGO algorithms. The wide applicability is one of the strong point\nnew{s} of the proposed technique.

The experimental results demonstrated the effect of the sample reuse in the compact GA and the pure rank-$\mu$ update CMA-ES. The effect of the sample reuse was visible up to $K \approx 5$ regardless of the problem dimension. \nnew{In the experiment of the CMA-ES}, the performance improvement by the sample reuse was emphasized when the population size was relatively small. We also introduced the rank-one update to the CMA-ES, which results in deviating from the IGO framework, and observed that the sample reuse can be combined with other techniques such as the rank-one update and reduces the number of function evaluations. A defect is observed with large values of $K$, the number of iterations to reuse populations, when applying the method to multimodal functions such as the Rastrigin function. Overall, a reasonable choice for $K$ seems to be $K \leq 3$. If the sample reuse is applied to other instances of the IGO algorithm, however, a reasonable choice of $K$ may be greater and will likely depend on other parameters, such as the population size and the learning rate. 

We note that a step-size adaptation is not introduced to the CMA-ES in this paper. In the CMA-ES, a step-size adaptation improves the efficiency and the robustness of the algorithm drastically. We barely observed any effect of the sample reuse in the CMA-ES with the step-size adaptation. The reason is that since the step-size adaptation leads to a fast change of the distribution, the likelihood ratio for the past populations will be very small, and the past population cannot contribute to the parameter update. We would like to emphasize that the motivation of this paper is not to improve well-developed algorithms such as the CMA-ES but to improve future algorithms derived from the IGO algorithms, e.g., the IGO algorithm on categorical variables or a mix of different types of variables. 

The idea of importance sampling may be useful not only to reuse past samples but to inject solutions that are not drawn from the current distribution. An example scenario is when we would like to inject an external solution, such as an optimal solution of a surrogate model, a best-so-far solution, repaired solutions, and so on  \citep{Hansen2011inria}. Another scenario is when the algorithm is implemented asynchronously, i.e., the parameter update is done immediately after receiving a pair of a solution and its objective value \citep{Glasmachers2013}. Both cases are rather important scenarios in practice, and the application of the importance sampling to these cases is a possible direction of future work.


\small
\bibliographystyle{apalike}
\bibliography{mycollection}

\appendix

\section{Proof: Variance Inequality}
\label{apdx:proof}

Let
\begin{align*}
  U(\theta^{(t)}) = \int g(x) \pthetat(x) dx
  = \frac{1}{K+1} \sum_{j=t-K}^{t} \int g(x) \frac{\pthetat(x)}{\pthetaj(x)} \pthetaj(x) \rmd x 
\end{align*}
be the quantity that we would like to estimate. We consider the following two estimators
\begin{align*}
  \tilde{U}_{\mathrm{IS1}}(\theta^{(t)})
  &:=  \sum_{j=t-K}^{t}  \frac{1}{n_j} \sum_{i=1}^{n_j}g(x^j_i) \frac{ c_j \pthetat(x^j_i)}{\pthetaj(x^j_i)}
  \\
  \tilde{U}_{\mathrm{IS2}}(\theta^{(t)})
  &:=  \sum_{j=t-K}^{t} \frac{1}{n_j} \sum_{i=1}^{n_j} g(x^j_i) \frac{c_j \pthetat(x^j_i)}{\sum_{k=t-K}^{t} c_k \pthetak(x^j_i)}
    \enspace,
\end{align*}
where $c_j$ satisfies $\sum_{j=t-K}^{t} c_j = 1$ and $c_j \geq 0$. Unbiasedness of both estimators are guaranteed by \cite[Section 3.2]{Veach1995}. Provided that $c_j / n_j = \alpha$ for all $j$, we will prove that $\Var[\tilde{U}_{\mathrm{IS2}}(\theta^{(t)})] \leq \Var[\tilde{U}_{\mathrm{IS1}}(\theta^{(t)})]$ for one dimensional $g$ . Then, we will generalize it to a multi-dimentional case.

The variances of both $\tilde{U}_{\mathrm{IS1}}(\theta^{(t)})$ and $\tilde{U}_{\mathrm{IS2}}(\theta^{(t)})$ are
\begin{align*}
  \Var[\tilde{U}_{\mathrm{IS1}}(\theta^{(t)})]
  &= \sum_{j=t-K}^{t}  \frac{c_j^2 }{n_j} \Var\left[g(x^j) \frac{ \pthetat(x^j)}{\pthetaj(x^j)} \right]
  \\
  \Var[\tilde{U}_{\mathrm{IS2}}(\theta^{(t)})]
  &=  \sum_{j=t-K}^{t} \frac{c_j^2}{n_j} \Var\left[ g(x^j) \frac{\pthetat(x^j)}{\sum_{k=t-K}^{t} c_k \pthetak(x^j)}\right]
    \enspace,
\end{align*}
respectively. Given $c_j / n_j = \alpha$, using Jensen's inequality $\sum_{j=t-K}^{t} \frac{c_j}{\pthetaj(x)} \geq \frac{1}{\sum_{k=t-K}^{t} c_k \pthetak(x)}$, we have
\begin{align*}
  \MoveEqLeft[2]\Var[\tilde{U}_{\mathrm{IS1}}(\theta^{(t)})] - \Var[\tilde{U}_{\mathrm{IS2}}(\theta^{(t)})]
  \\
  &= \sum_{j=t-K}^{t} \frac{c_j^2}{n_j} \int g(x)^2 \pthetat(x)^2 \left(\frac{1}{\pthetaj(x)} - \frac{1}{\sum_{k=t-K}^{t} c_k \pthetak(x)}\right) \mathrm{d} x
    \\
  &= \alpha\int g(x)^2 \pthetat(x)^2 \left(\sum_{j=t-K}^{t} \frac{c_j}{\pthetaj(x)} - \frac{1}{\sum_{k=t-K}^{t} c_k \pthetak(x)}\right) \mathrm{d} x
    \geq 0\enspace.
\end{align*}
Hence, $\Var[\tilde{U}_{\mathrm{IS1}}(\theta^{(t)})] - \Var[\tilde{U}_{\mathrm{IS2}}(\theta^{(t)})] \succcurlyeq 0$. 
Similarly, if $g$ takes the value in $\R^d$, the covariance matrices of these estimators \del{satisfies}{}\nnew{satisfy}
\begin{align*}
  \MoveEqLeft[2]\Cov[\tilde{U}_{\mathrm{IS1}}(\theta^{(t)})] - \Cov[\tilde{U}_{\mathrm{IS2}}(\theta^{(t)})]
  \\
  &= \sum_{j=t-k}^{t} \frac{c_j^2}{n_j} \int g(x)g(x)^\T \pthetat(x)^2 \left(\frac{1}{\pthetaj(x)} - \frac{1}{\sum_{k=t-K}^{t} c_k \pthetak(x)}\right) \mathrm{d} x
    \\
  &= \alpha\int g(x)g(x)^\T \pthetat(x)^2 \left(\sum_{j=t-K}^{t} \frac{c_j}{\pthetaj(x)} - \frac{1}{\sum_{k=t-K}^{t} c_k \pthetak(x)}\right) \mathrm{d} x
    \succcurlyeq 0\enspace,
\end{align*}
where $A \succcurlyeq 0$ for a symmetric $A$ means $x^\T A x \geq 0$ for any nonzero $x \in \R^d$, i.e., all the eigenvalues of $A$ are nonnegative. Hence, $\Cov[\tilde{U}_{\mathrm{IS1}}(\theta^{(t)})] - \Cov[\tilde{U}_{\mathrm{IS2}}(\theta^{(t)})] \succcurlyeq 0$. 

\end{document}